\documentclass[11pt,letterpaper]{article}
\usepackage[letterpaper]{geometry}
\usepackage{acl2012}
\usepackage{times}
\usepackage{latexsym}
\usepackage{amsmath}
\usepackage{multirow}
\usepackage{url}
\makeatletter
\newcommand{\@BIBLABEL}{\@emptybiblabel}
\newcommand{\@emptybiblabel}[1]{}
\makeatother

\usepackage{amsmath,amssymb}
\usepackage{graphicx}
\usepackage{multirow}
\usepackage{booktabs}
\usepackage{hyperref} 
\usepackage{mathtools}
\usepackage{subcaption}
\usepackage{flushend}
\usepackage{algorithm}
\usepackage{algpseudocode}

\title{Joint Modeling of Topics, Citations, and Topical Authority in Academic Corpora}

\author{Jooyeon Kim \\
  KAIST \\
  {\tt \small{jooyeon.kim@kaist.ac.kr}} \And
  Dongwoo Kim \\
  Australian National University \\
  {\tt \small{dongwoo.kim@anu.edu.au}} \And
  Alice Oh \\  
  KAIST \\
  {\tt \small{alice.oh@kaist.edu}} }

\date{}

\begin{document}
\maketitle

\begin{abstract}

Much of scientific progress stems from previously published findings, but searching through the vast sea of scientific publications is difficult. We often rely on metrics of scholarly authority to find the prominent authors but these authority indices do not differentiate authority based on research topics.  We present Latent Topical-Authority Indexing (LTAI) for jointly modeling the topics, citations, and topical authority in a corpus of academic papers. Compared to previous models, LTAI differs in two main aspects. First, it explicitly models the generative process of the citations, rather than treating the citations as given. Second, it models each author's influence on citations of a paper based on the topics of the cited papers, as well as the citing papers. 
We fit LTAI to four academic corpora: CORA, Arxiv Physics, PNAS, and Citeseer. We compare the performance of LTAI against various baselines, starting with the latent Dirichlet allocation, to the more advanced models including author-link topic model and dynamic author citation topic model. The results show that LTAI achieves improved accuracy over other similar models when predicting words, citations and authors of publications. 

\end{abstract}

\section{Introduction}

\begin{figure}[t!]
	\centering
 	\includegraphics[width=0.85\linewidth]{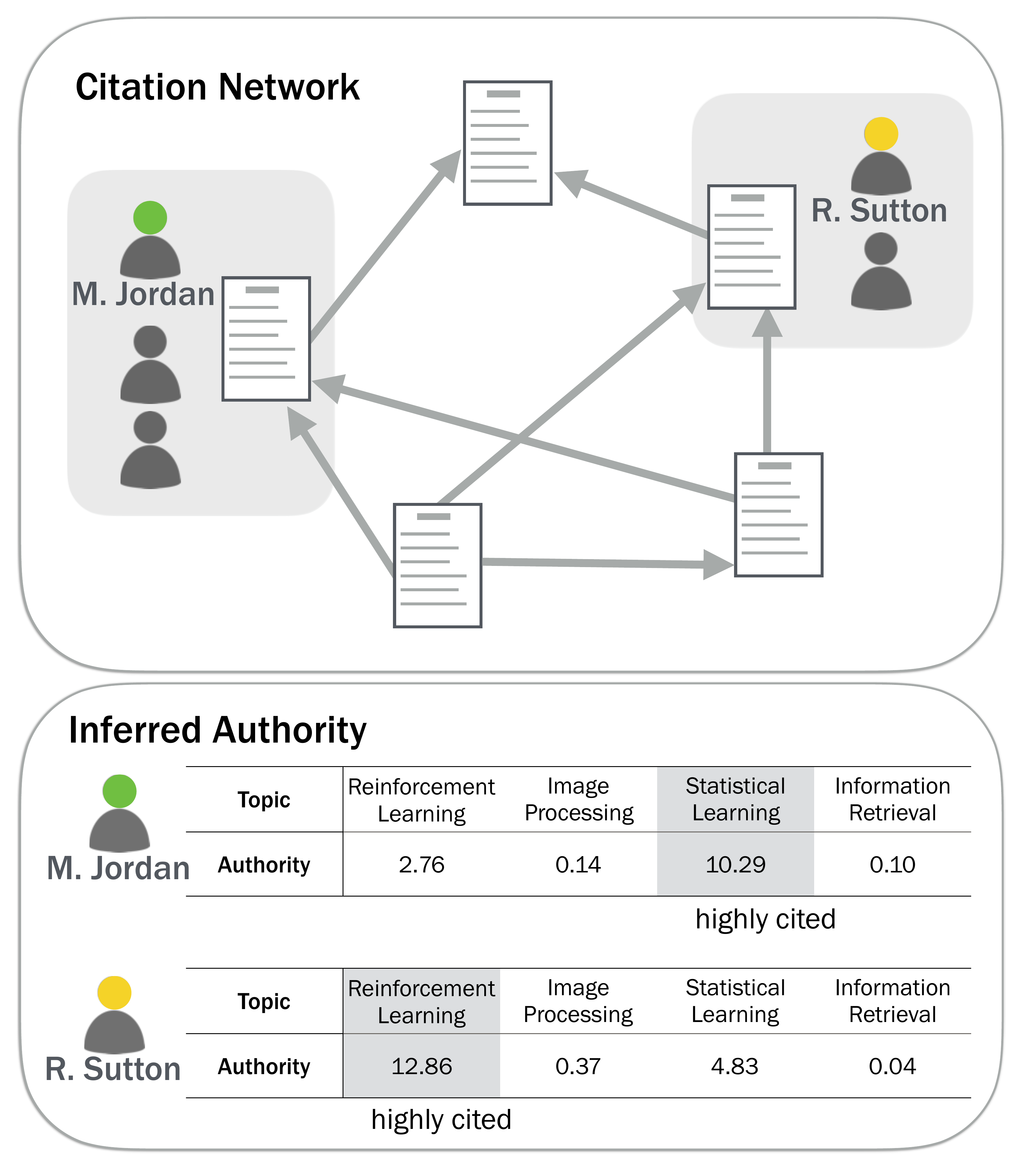}	
	\caption{\label{fig:intro}Overview of Latent Topical Authority Indexing (LTAI). Based on content, citation, and authorship information (top), the LTAI discovers topical authority of authors; it increases when a paper with certain topics gets cited (bottom). Topical authority examples are the results of the LTAI with CORA dataset and 100 topics.}
\end{figure}

With a corpus of scientific literature, we can observe the complex and intricate process of scientific progress. We can learn the major topics in journal articles and conference proceedings, follow authors who are prolific and influential, and find papers that are highly cited. The huge number of publications and authors, however, makes it practically impossible to attain any deep or detailed understanding beyond the very broad trends. For example, if we want to identify authors who are particularly influential in a specific research field, it is difficult to do so without the aid of automatic analysis.

Online publication archives, such as Google Scholar, provide near real-time metrics of scholarly impact, such as the h-index \cite{hirsch2005index}, the journal impact factor \cite{garfield2006history}, and citation count. Those indices, however, are still at a coarse level of granularity. For example, both Michael Jordan and Richard Sutton are researchers with very high citation count and h-index, but they are authoritative in different topics, Jordan in the more general machine learning topic of statistical learning, and Sutton in the topic of reinforcement learning. It would be much more helpful to know that via topical authority scores, as shown in Figure~\ref{fig:intro}.

Fortunately, various academic publication archives contain the full contents, references, and meta-data including titles, venues, and authors. With such data, we can build and fit a model to partition researchers' scholarly domain into topics at a much finer-grain and discover their academic authority within each topic. To do that, we propose a model named Latent Topical-Authority Indexing (LTAI), based on the latent Dirichlet allocation, to jointly model the topics, authors' topical authority, and citations among the publications. 

We illustrate the modeling power of the LTAI with four corpora encompassing a diverse set of academic fields: CORA, Arxiv Physics, PNAS, and Citeseer. To show the improvements over other related models, we carry out prediction tasks on word, citation and authorship using the LTAI and compare the results with those of latent Dirichlet allocation \cite{Blei:2003p4796}, relational topic model \cite{chang2010hierarchical}, author-link topic model, and dynamic author-cite topic model \cite{kataria2011context}, as well as simple baselines of topical h-index. The results show that the LTAI outperforms these other models for all prediction tasks.

The rest of this paper is organized as follows. In~\autoref{sec:related}, we describe related work, including models that are most similar to the LTAI, and describe how the LTAI fits in and contributes to the field. In~\autoref{sec:model}, we describe the LTAI model in detail and present the generative process. In~\autoref{sec:inference}, we explain the algorithm for approximate inference, and in~\autoref{sec:opt}, we present a faster algorithm for scalability. In~\autoref{sec:exp}, we describe the experimental setup and in~\autoref{sec:eval}, we present the results to show that the LTAI performs better than other related models for word, citation and authorship prediction.


\section{\label{sec:related}Related Work}


In this section, we review related papers, first in the field of NLP and ML-based analysis of scientific corpora, then the approaches based on the Bayesian topic models for academic corpora, and lastly joint models of topics, authors, and citations. In analyzing scientific corpora, previous research presents classifying scientific publications \cite{caragea2015co}, recommending yet unlinked citations \cite{huang2015neural,Neiswanger2014modeling,wang2015ldtm,jiang2015chronological}, summarizing and extracting key phrases \cite{cohan2015scientific,caragea2014citation}, triggering better model fit \cite{he2015discovering}, incorporating authorship information to increase the content and link predictability \cite{sim2015utility}, estimating a paper's potential influence on academic community \cite{dong2015will}, and finding and classifying different functionalities of citation practices \cite{moravcsik1975some,teufel2006automatic,valenzuela2015identifying}.

Several variants of topic modeling consider the relationship between topics and citations in academic corpora. Topic models that use text and citation network are divided into two types: (a) models that generate text given citation network \cite{Dietz:2007:UPC:1273496.1273526,foulds2013modeling} and (b) models that generate citation network given text \cite{nallapati2008joint,liu2009topic,chang2010hierarchical}. While our model falls into the latter category, we also take into account the influence of the authors on the citation structure.




Most closely related to the LTAI are the citation author topic model \cite{tu2010citation}, the author-link topic model, and the dynamic author-cite topic model \cite{kataria2011context}. Similar to the LTAI, they are designed to capture the influence of the authors. However, these models infer authority by referencing only the citing papers' text, while our authority is based on the predictive modeling of comparing both the citing and the cited papers. Furthermore, the LTAI defines a generative model of citations and publications by introducing a latent authority index, whereas the previous models assume the citation structure is given. the LTAI thus explicitly gives a topical authority index, which directly answers the question of which author increases the probability of a paper being cited.
\section{Latent Topical-Authority Indexing} 

\label{sec:model}

The LTAI models the complex relationship among the topics of publications, the topical authority of the authors, and the citations among these publications. The generative process of the LTAI can be divided into two parts: content generation and citation network generation. We make several assumptions in the LTAI to model citation structure of academic corpora. First, we assume a citation is more likely to occur between two papers that are similar in their topic proportions. Second, we assume that an author differs in their authority (i.e., potential to induce citation) for each topic, and an author's topical authority positively correlates with the probability of citation among publications.  Also, in the LTAI, when there are multiple authors in a single cited publication, their contribution of forming citations with respect to different citing papers varies according to their topical authority. Lastly, we assign different concentration parameters for a pair of papers with and without citation. In this paper, we use terms positive and negative links to denote pairs of papers with and without citations respectively. 

\begin{figure}[t!]
	\centering
 	\includegraphics[width=0.95\linewidth]{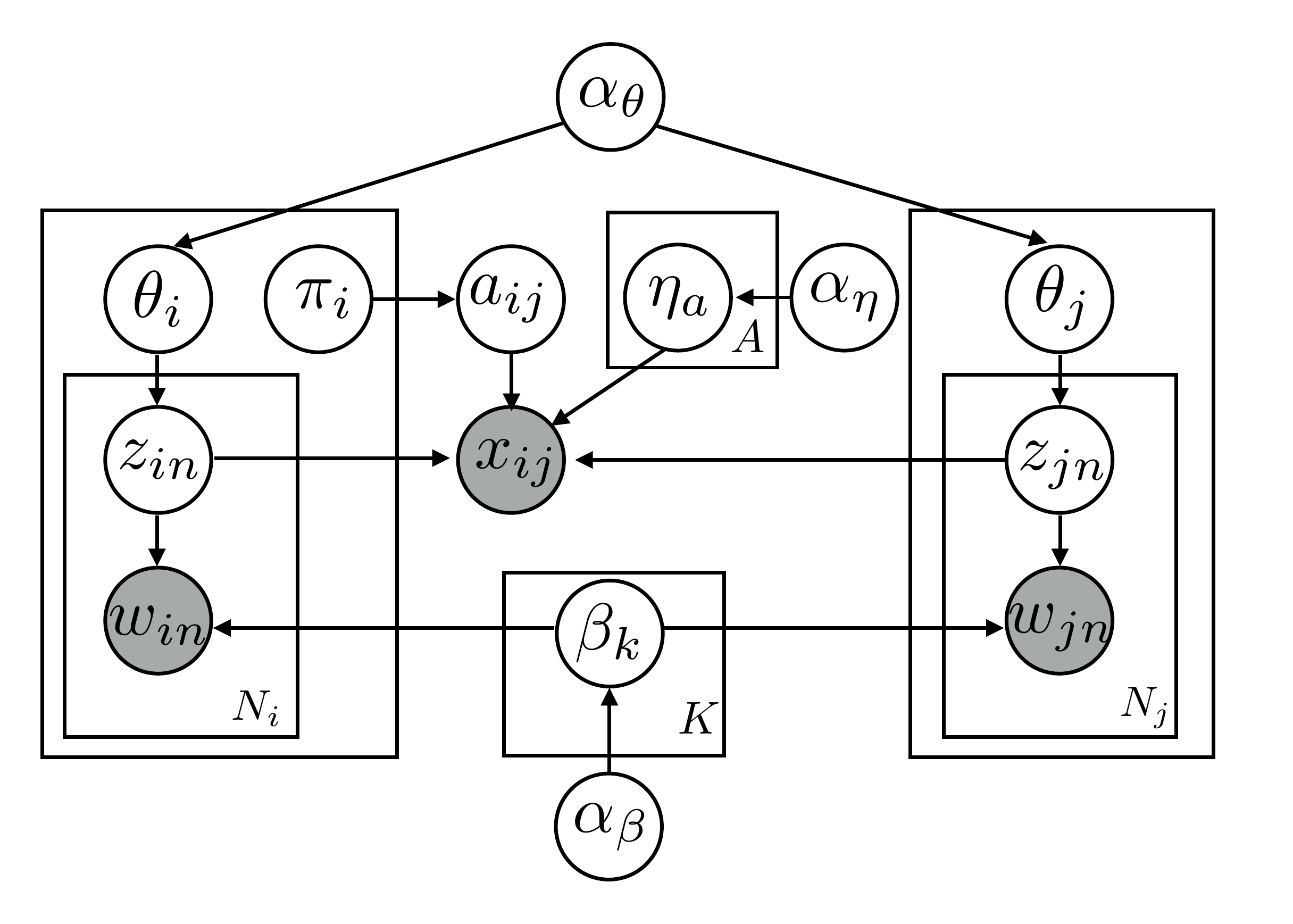}		
	\caption{\label{fig:graphmodel}Graphical representation of the LTAI. The LTAI jointly models content-related variables $\theta, z, w$, $\beta$, and author and citation related variables $\eta$ and $\pi$.}
\end{figure}

Figure \ref{fig:graphmodel} illustrates the graphical model of the LTAI, and we summarize the generative process of the LTAI, where the variables of the model are explained in the remainder of this section, as follows:
\begin{enumerate}
	\item For each topic $k$, draw topic $\beta_k \sim \text{Dir}(\alpha_\beta)$.
	\item For each document $i$:
	\begin{enumerate}
		\item Draw topic proportion $\theta_i \sim \text{Dir}(\alpha_\theta)$.
		\item For each word $w_{in}$:
		\begin{enumerate}
			\item Draw topic assignment $z_{in}|\theta_i \sim \text{Mult}(\theta_i)$.
			\item Draw word $w_{in}|z_{in}, \beta_{1:K} \sim \text{Mult}(\beta_{z_{in}})$.
		\end{enumerate}
	\end{enumerate}
	\item For each author $a$ and topic $k$:
		\begin{enumerate}
		\item Draw authority index $\eta_{ak} \sim \mathcal{N}(0, \alpha_\eta^{-1})$.
		\end{enumerate}
	\item For each document pair from $i$ to $j$:
		\begin{enumerate}
		\item Draw influence proportion \\ parameter $\pi_{i \leftarrow j} \sim \text{Dir}(\pi_i)$.
		\item Draw author $a_i | \pi_{i \leftarrow j} \sim \text{Mult}(\pi_{i \leftarrow j})$.
		\item Draw link $x_{i \leftarrow j} \sim \\  \mathcal{N}(\sum_k \eta_{a_{i}k} \bar{z}_{ik}\bar{z}_{jk}, c_{i\leftarrow j}^{-1})   $.
		\end{enumerate}
\end{enumerate}

\subsection{Content Generation}

To model the content of publications, we follow a standard document generative process of latent Dirichlet allocation (LDA) \cite{Blei:2003p4796}. Also, we inherit notations for variables from LDA; $\theta$ is the per-document topic distribution, $\beta$ is the per-topic word distribution, $z$ is the topic for each word in a document where $w$ is the corresponding word, and $\alpha_\theta$, $\alpha_\beta$ are the Dirichlet parameters of $\theta$ and $\beta$ respectively. 


\subsection{Citation Generation}
\label{subsec:citation_generation}

Let $x_{i \leftarrow j}$ be a binary valued variable which indicates that publication $j$ cites publication $i$. We formulate a continuous variable $r_{i \leftarrow j}$ which is a linear combination of the authority variable and the topic proportion variable to approximate $x_{i \leftarrow j}$ by minimizing the sum of squared errors between the two variables. There is a body of research on using continuous user and item-related variables to approximate binary variables in the field of recommender systems  \cite{rennie2005fast,koren2009matrix}.

Approximating binary variables using linear combination of continuous variables can be probabilistically generalized \cite{salakhutdinov2007probabilistic}. Using probabilistic matrix factorization, we approximate probability mass function $p(x_{i \leftarrow j})$ using probability density function $\mathcal{N}(x_{i \leftarrow j} | r_{i \leftarrow j}, c_{i\leftarrow j}^{-1})$, where the precision parameter $c_{i\leftarrow j}$ can be set differently for each pair of papers as will be discussed below. 

\textbf{Content Similarity Between Publications}: In the LTAI, we model relationship between a random pair of documents $i$ and $j$. The probability of publication $j$ citing publication $i$ is proportional to the similarity of topic proportions of two publications, i.e., $r_{i  \leftarrow j}$ positively correlates to $\sum_k \theta_{ik} \theta_{jk}$. Following relational topic model's approach \cite{chang2010hierarchical}, we use $\bar{z}_i = \frac{1}{N_{i}} \sum_{n}{z_{i,n}} \approx \theta_i$ instead of topic proportion parameter $\theta_i$.

\textbf{Topical Authority of Cited Paper}: We introduce a $K$-dimensional vector $\eta_{a}$ for representing the topical authority index of author $a$. $\eta_{ak}$ is a real number drawn from the zero-mean normal distribution with variance $\alpha_\eta^{-1}$. Given the authority indices $\eta_{a_i}$ for author $a$ of cited publication $i$, the probability of citation is further modeled as  $r_{i  \leftarrow j} = \sum_k \eta_{a_{i}k}\bar{z}_{ik} \bar{z}_{jk}$, where the authority indices can promote or demote the probability of citation. 

\textbf{Different Degree of Contribution among Multiple Authors}: Academic publications are often written by more than one author. Thus, we need to distinguish the influence of each author on a citation between two publications. Let $\mathcal{A}_i$ be a set of authors of publication $i$. To measure the influence proportion of author $a \in \mathcal{A}_i$ on the citation from $i$ to $j$, we introduce additional parameter $\pi_{ij}$ which is a one-hot vector drawn from a Dirichlet distribution with $|\mathcal{A}_i|$-dimensional parameter $\pi_i$. $\pi_{i \leftarrow j a} \in \{0, 1 \}$ is an element of $\pi_{i \leftarrow j}$ which measures the influence of author $a$ on the citation from $j$ to $i$ and sums up to one ($\sum_{a \in \mathcal{A}_{i}} \pi_{i \leftarrow ja} = 1$) over all authors of publication $i$. 
We approximate the probability of citation $x_{i  \leftarrow j}$ from publication $j$ to publication $i$ by $\label{eqn:link_prob}p(x_{i \leftarrow j} | z, \pi_{ij}, a_{i \leftarrow j}, \eta_a ) \approx \sum_{a \in \mathcal{A}_i} \pi_{i \leftarrow ja} \mathcal{N}(x_{i \leftarrow  j} | \sum_k \eta_{ak} \bar{z}_{ik}\bar{z}_{jk}, c_{i \leftarrow j}^{-1})$ which is a mixture of normal distributions with precision parameter $c_{i \leftarrow j}$. Therefore, if topic distributions of paper $i$ and $j$ are similar and if $\eta$ values of the cited paper's authors are high, the citation formation probability increases; on the other hand, dissimilar or topically irrelevant pair of papers with less authoritative authors on the cited paper will be assigned with low probability of citation formation.  

\textbf{Different Treatment between Positive and Negative links}: Citation is a binary problem where $x_{i \leftarrow j}$ is either one or zero. When $x_{i \leftarrow j}$ is zero, this can be interpreted in two ways: 1) the authors of citing publication $j$ are unaware of the publication $i$, or 2) the publication $j$ is not relevant to publication $i$. Identifying which case is true is impossible unless we are the authors of the publication. Therefore the model embraces this uncertainty in the absence of a link between publications. 
We control the ambiguity by the Gaussian distribution with precision parameter $c_{i   j}$ as follows:
\begin{align}
c_{i \leftarrow j} = \left\{ 
  \begin{array}{l l}
    c_+ & \quad \text{if $x_{i \leftarrow  j} = 1$}\\
    c_- & \quad \text{if $x_{i  \leftarrow j} = 0$}
  \end{array} \right.
\end{align}
where $c_+ > c_-$ to ensure that we have more confidence on the observed citations. This is an implicit feedback approach that permits using negative examples ($x_{i \leftarrow  j} = 0$) of sparse observations by mitigating their importance \cite{hu2008collaborative,wang2011collaborative,purushotham2012collaborative}. Setting different values to the precision parameter $c_{i \leftarrow j}$ according to $x_{i \leftarrow j}$ induces cyclic dependencies between the two variables, and due to this cycle, the model no longer becomes a Bayesian network, or a directed acyclic graph. However, we note that this setting does lead to better experimental results, and we show the pragmatic benefit of the setting in the Evaluation section.

\subsection{Joint Modeling of the LTAI}

In the LTAI, the topics and the link structures are simultaneously learned, and thus the content-related variables and the citation-related variables mutually reshape one another during the posterior inference. On the other hand, if content and citation data are modeled separately, the topics would not reflect any information about the document citation structure. Thus, in the LTAI, documents with shared links are more likely to have similar topic distributions which leads to better model fit. We develop and explain this joint inference in~\autoref{sec:inference}. In~\autoref{sec:eval}, we illustrate the differences in word-level predictive powers of the LTAI and LDA. 


\section{Posterior Inference}

\label{sec:inference}
We develop a hybrid inference algorithm in which the posterior of content-related parameters $\theta$, $z$, and $\beta$ are approximated by variational inference, and author-related parameters $\pi$ and $\eta$ are approximated by EM. In algorithm \ref{alg:ltai}, we summarize the full inference procedure of the LTAI.

\subsection{Content Parameters: Variational Update}
Since computing the posterior distribution of the LTAI is intractable, we use variational inference to optimize variational parameters each of which correspond to original content-related variables. Following the standard mean-field variational approach, we define fully factorized variational distributions over the topic-related latent variables $q(\theta, \beta, z) = \prod_{i} q(\theta_i|\Psi_{in}) \prod_{N_i} q(z_{in}|\gamma_{i}) \prod_{k} q(\beta_k|\lambda_{k})$ where for each factorized variational distribution, we place the same family of distributions as the original distribution.
Using the variational distributions, we bound the log-likelihood of the model as follows:
\begin{align}
\mathcal{L}_{[q]} &= \mathbb{E}_q[\sum_k \log p(\beta_k | \alpha_\beta) + \sum_i \log p(\theta_i | \alpha_\theta)  \notag\\
&+ \sum_i \sum_{N_i} \log p(z_{in} | \theta_d) + \log p(w_{in} | \beta_{z_{in}}) 	\\
&+  \sum_{i,j} \log p(x_{i \leftarrow j} | z_{i}, z_{j}, \pi_{i})] - \mathcal{H}[q] \notag
\end{align}
where $\mathcal{H}[q]$ is the negative entropy of $q$. 

Taking the derivatives of this lower bound with respect to each variational parameter, we can obtain the coordinate ascent updates. The update for the variational Dirichlet parameters $\gamma_{i}$ and the $\lambda_{k}$ is the same as the standard variational update for LDA \cite{Blei:2003p4796}.
The update for the variational multinomial $\phi_{in}$ is:
\begin{align}
\label{eqn:phi0}
\phi_{ink} &\propto \exp\left\{\frac{\sum_{j} \partial \mathbb{E}_q[ \log p(x_{i \leftarrow j} | \bar{z_{i}}, \bar{z_{j}}, \pi_i, \eta) ]}{\partial{\phi_{ink}}} \right.  \notag \\
 &+ \frac{\sum_{j} \partial \mathbb{E}_q[ \log p(x_{j \leftarrow i } | \bar{z_{j}}, \bar{z_{i}}, \pi_j, \eta) ]}{\partial{\phi_{ink}}} & \\
& + \left. \mathbb{E}_q[\log \theta_{ik}] + \mathbb{E}_q[\log \beta_{kw_{in}}] \right\} \notag &
\end{align}
where the gradient of expected log probabilities of both incoming link $x_{i \leftarrow j}$ and outgoing link $x_{j \leftarrow i }$ contribute to the variational parameter. The first expectation can be rewritten as
\begin{align}
\label{eqn:phi}
&\mathbb{E}_q[ \log p(x_{i \leftarrow j} | \bar{z_{i}}, \bar{z_{j}}, \pi_i, \eta) ] &\\
&\quad = \mathbb{E}_q[ \log \sum_{a\in A_i} p(a_{i \leftarrow j}=a|\pi_i) p( x_{i \leftarrow j} | \bar{z_{i}}, \bar{z_{j}}, \eta_{a})] & \notag \\  
&\quad \geq  \sum_{a\in A_i} p(a_{i \leftarrow j}=a|\pi_i) \mathbb{E}_q[ \log p( x_{i \leftarrow j} | \bar{z_{i}}, \bar{z_{j}}, \eta_{a})] \notag &
\end{align}
where $A_i$ is the set of authors of $i$. We take the lower bound of the expectation using Jensen's inequality. The last term is approximated by the first order Taylor expansion $\mathbb{E}_q[ \log p( x_{i \leftarrow j} | \bar{z_{i}}, \bar{z_{j}}, \eta_a)] = \mathcal{N}(x_{i \leftarrow j} | \bar{\phi}_i ^{\top} \text{diag}(\eta_a) \bar{\phi}_j, c_{i \leftarrow j}^{-1})$.
Finally, the approximated gradient of $\phi_{ink}$ with respect to the incoming directions to document $i$ is
{\small
\begin{align}
\label{eqn:phi2}
&\frac{\sum_{j} \partial \mathbb{E}_q[ \log p(x_{i \leftarrow j} | \bar{z_{i}}, \bar{z_{j}}, \pi_i, \eta) ]}{\partial{\phi_{ink}}}  \approx \\
& \sum_j \frac{\bar{\phi}_{jk} c_{i \leftarrow j}}{N_i}  \sum_{a \in A_i} \eta_{ak} (x_{i \leftarrow j} - \bar{\phi}_i^{\top} \text{diag}(\eta_a) \bar{\phi}_j) p(a|\pi_i)  \notag
\end{align}}
where diag is a diagonalization operator and $\bar{\phi_i}$ is $\sum_{n=1}^{N_i} \phi_{in}/N_i$.
We can compute the gradient with respect to the outgoing directions in the same way.

\begin{algorithm}[t!]
\caption{Posterior inference algorithm for the LTAI}
\label{alg:ltai}
\begin{algorithmic}
\State Initialize $\gamma$, $\lambda$, $\pi$, and $\eta$ randomly 
\State Set learning-rate parameter $\rho_t$ that satisfies \\ Robbins-Monro condition
\State Set subsample sizes $S_V$, $S_E$, $S_S$ and $S_A$
\Repeat

\State \small{\textit{Variational update: local publication parameters}}
\State $\mathcal{S}_S \leftarrow$ $S_S$ randomly sampled publications
\For{ $i$ in $\mathcal{S}_S$}
\For{ $n=1$ to $N_i$}
\State \parbox[t]{\dimexpr\linewidth-\algorithmicindent}{$S_{\leftarrow}, S_{\rightarrow} \gets$ Set of $S_V$ random samples \strut}
\State {Update $\phi_{ink}$ using Equation \ref{eqn:phi},  \ref{eqn:phi2},  \ref{eqn:phi3}.}
\EndFor
\State $\gamma_{i} \gets \alpha_\theta + \sum_{N_i} \phi_{in}$
\EndFor

\State

\State \small{\textit{EM update: local author parameters}}
\State $\mathcal{S}_A \leftarrow S_A$ randomly sampled authors
\For{ $a$ in $\mathcal{S}_A$}
\State $\mathcal{S}_E \leftarrow S_E$ random publication pairs
\State Update $\eta_a$ using Equation \ref{eqn:eta}, \ref{eqn:eta_stochastic}
\For{$i$ in $\mathcal{D}_a$ and $j = 1$ to $D$}
\State $\pi_{i \leftarrow j a} \propto \pi_{ia} \mathcal{N}(\bar{z_{i}}^{\top} \eta_a \bar{z_{j}}, c_{i \leftarrow j}^{-1})$
\EndFor
\EndFor

\State

\State \small{\textit{Stochastic variational update}}
\For{ $k=1$ to $K$}
\State $\hat{\lambda}_k \leftarrow \alpha_\beta + \frac{D}{S_S}\sum^{\mathcal{S}_S}_{d=1}{  \sum^{N_d}_{n=1}{\phi^{k}_{dn} w_{dn}  } }$
\EndFor

\State Set $\lambda^{(t)} \leftarrow (1-\rho^t)\lambda^{(t-1)} + \rho_t \hat{\lambda}$

\Until{satisfying converge criteria}

\end{algorithmic}
\end{algorithm}

\subsection{Author Parameters: EM Step}
\label{emstep}

We use the EM algorithm to update author-related parameters $\pi$, and $\eta$ based on the lower bound computed by variational inference. In the E step, we compute the probability of author contribution to the link between document $i$ and $j$. 
\begin{align}
{\small \pi_{i \leftarrow j a} = \frac{ \pi_{ia} \mathcal{N}(\bar{z_{i}}^{\top} \eta_a \bar{z_{j}}, c_{i \leftarrow j}^{-1}) }{ \sum\limits_{a' \in A_i}^{} \pi_{ia'} \mathcal{N}(\bar{z_{i}}^{\top} \eta_{a'} \bar{z_{j}}, c_{i \leftarrow j}^{-1})}}
\end{align}

In the M step, we optimize the authority parameter $\eta$ for each author. Given the other estimated parameters, taking the gradient of $\mathcal{L}$ with respect to $\eta_a$ and setting it to zero leads to the following update equation:
\begin{align}
\label{eqn:eta}
\eta_{a} = (\Psi^{\top}_{a} C_{a} \Psi_{a} + \alpha_{\eta}I)^{-1} \Psi^{\top}_{a} C_{a} X_{a}
\end{align}

Let $\mathcal{D}_a$ be the set of documents written by author $a$ and $\mathcal{D}_{a}(i)$ be the $i$th document written by $a$. Then $\Psi_{a}$ is a vertical stack of $|\mathcal{D}_a|$ matrices $\Psi_{\mathcal{D}_{a}(i)}$, whose $j$th row is $ \bar{\phi}_{\mathcal{D}_{a}(i)} \circ \bar{\phi}_{j}$, the Hadamard product between $\bar{\phi}_{\mathcal{D}_{a}(i)}$ and $\bar{\phi}_{j}$. Similarly, $C_{a}$ is a vertical stack of  $|\mathcal{D}_a|$ matrices $C_{\mathcal{D}_{a}(i)}$ whose $j$ th diagonal element is $c_{\mathcal{D}_{a}(i) \leftarrow j}$, and $X_{a}$ is a vertical stack of $|\mathcal{D}_a|$ vectors $X_{\mathcal{D}_{a}(i)}$ whose $j$ th element is $\pi_{\mathcal{D}_{a}(i) \leftarrow ja} \times x_{\mathcal{D}_{a}(i) \leftarrow j}$. Finally, we update $\pi_{  \mathcal{D}_{a}(i)  a} = \sum_{j}{\pi_{  \mathcal{D}_{a}(i) \leftarrow ja}}/{D}$.

\section{Faster Inference Using Stochastic Optimization}
\label{sec:opt}

To model topical authority, the LTAI considers the linkage information. If two papers are linked by citation, the topical authority of the cited paper's authors will increase while the negative link buffers the potential noise of irrelevant topics. This algorithmic design of the LTAI results in high model complexity. To remedy this issue, we adopt the noisy gradient method from the stochastic approximation algorithm \cite{robbins1951stochastic} to subsample negative links for updating per-document topic variational parameter $\phi$ and authority parameter $\eta$. The prior work of using subsampled negative links to reduce computational complexity is introduced in \cite{raftery2012fast}. Also, we elucidate how stochastic variational inference \cite{hoffman2013stochastic} is applied in our model to update global per-topic-word variational parameter $\lambda$. 

\subsection{Updating $\phi$ and $\eta$}
Updating $\bar{\phi}_{i}$ for document $i$ in variational update requires iterating over every other document and computing the gradient of link probability. This leads to the time complexity $\mathcal{O}(DK)$ for every $\bar{\phi}_{i}$. 

To apply the noisy gradient method, we divide the gradient of the expected log probability of link into two parts:
{\small
\begin{align}
&\sum_{j} \frac{ \partial \mathbb{E}_q[ \log p(x_{i \leftarrow j} | \bar{z_{i}}, \bar{z_{j}}, \pi_i) ]}{\partial{\phi_{ink}}} =& \\
& \sum_{j : x_{i \leftarrow j} = 1} \frac{ \partial \mathbb{E}_q[ \log p(x_{i \leftarrow j} ) ]}{\partial{\phi_{ink}}} 
+ \sum_{j : x_{i \leftarrow j} = 0} \frac{ \partial \mathbb{E}_q[ \log p(x_{i \leftarrow j} ) ]}{\partial{\phi_{ink}}} & \notag
\end{align}}%
where the first and the second term of RHS is the gradient sum of positive links $(x_{ij}=1)$ and negative links ($x_{ij}=0$), respectively. Compared to positive links, the order of negative links is close to the total number of documents, and thus computing the second term results in computational inefficiency. However, in our model, we reduced the importance of the negative links by assigning a larger variance $c_{ij}^{-1}$ compared to the positive links, and the empirical mean of $\bar\phi_{j}$ for negative links follows the Dirichlet expectation due to the large number of negative links. Therefore, we approximate the expectation of the gradient for the negative links using the noisy gradient as follows:
{\small
\begin{align}
\label{eqn:phi3}
\sum_{j : x_{i \leftarrow j} = 0} \frac{ \partial \mathbb{E}_q[ \log p(x_{i \leftarrow j} ) ]}{\partial{\phi_{ink}}} = \frac{D^{-}_{i}}{S_{V}} \sum_{\mathcal{S}_V} \frac{ \partial \mathbb{E}_q[ \log p(x_{i \leftarrow s} ) ]}{\partial{\phi_{ink}}}
\end{align}}%
where $D^{-}_{i}$ is the number of negative links (i.e. $x_{i \leftarrow j}=0$) of document $i$, and $S_{V}$ is the size of sub-samples $\mathcal{S}_V$ for the variational update. We randomly sample $S_V$ documents, compute gradients on the sampled documents, and then scale the average gradient to the size of the negative link $D^{-}_{i}$. This noisy gradient method reduces the updating time complexity from $\mathcal{O}(DK)$ to $\mathcal{O}(S_{V}K)$.

Now, we discuss how to approximate author's topical authority based on \autoref{eqn:eta}. When $K \ll D \times D_{a}$, the computational bottleneck is $\Psi^{\top}_{a} C_{a} \Psi_{a}$ which has time complexity $\mathcal{O}(DD_{a}K^{2})$. To alleviate this complexity, we once again approximate the large number of negative links using smaller number of subsamples. Specifically, while keeping the positive link rows $\Psi_{a^+}$ intact, we approximate negative link rows in $\Psi_a$ using smaller matrix $\Psi_{a^-}$ that has $S_E$ rows, or the size of subsamples for the EM step. Using this approximation, we can represent $\Psi^{\top}_{a} C_{a} \Psi_{a}$ as

\begin{align}
\label{eqn:eta_stochastic}
\Psi^{\top}_{a} C_{a} \Psi_{a} =  c_{+} \Psi^{\top}_{a^+} \Psi_{a^+} + \frac{c_{-}  D^{-}_{a}}{S_E} \Psi^{\top}_{a^-} \Psi_{a^-}
\end{align}
with the time complexity of $\mathcal{O}(S_{E}K^2)$, where $D^{-}_{a}$ is the number of rows with negative links in $\Psi_{a}$. Also, although we do not incorporate rigorous analysis on the performance of our model given the size of the subsamples, we confirm that the negative link size greater than 100 does not degrade the model performance in any of our experiment. 

\begin{figure}[t!]
	\centering
 	\includegraphics[width=0.6\linewidth]{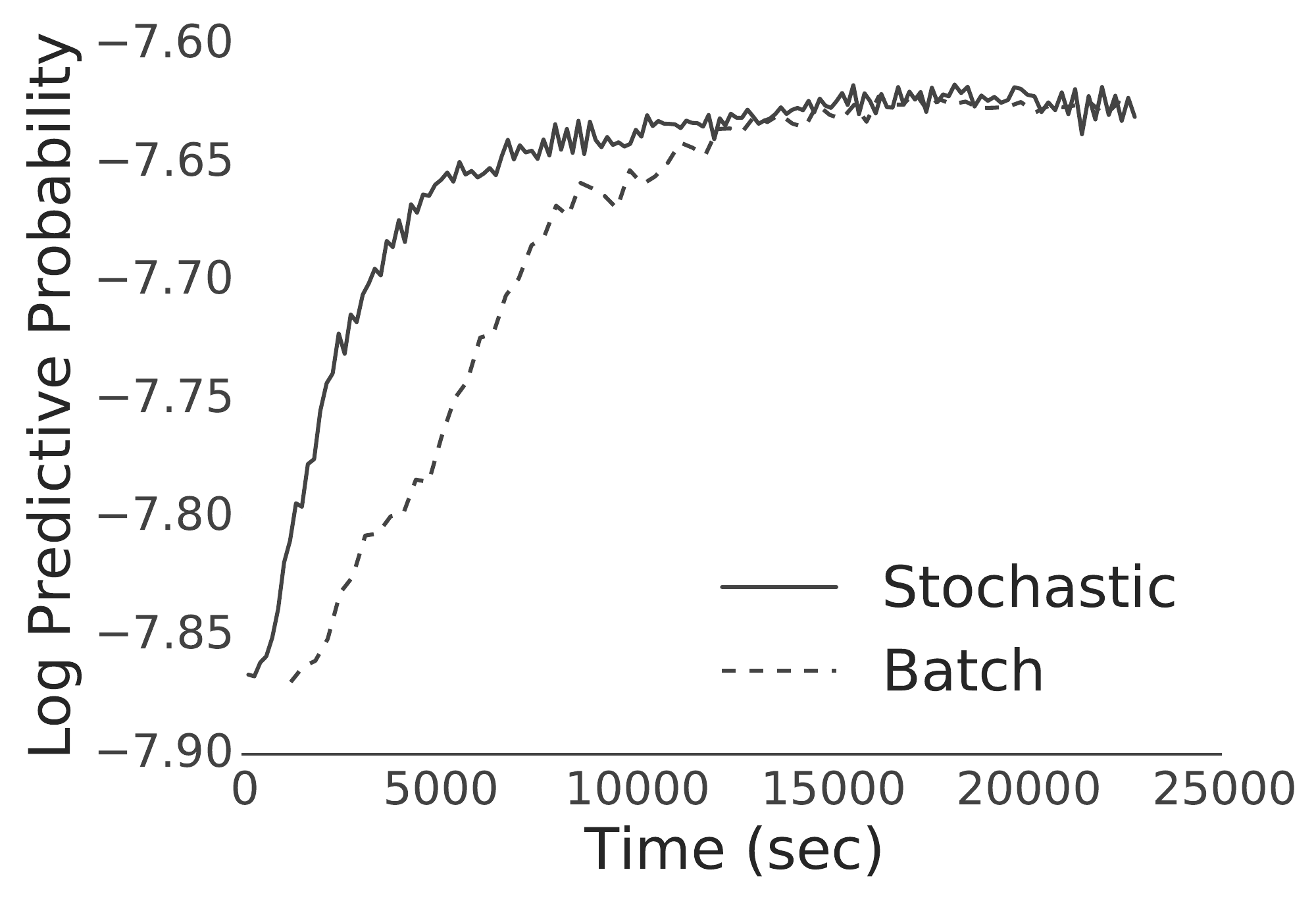}		
	\caption{\label{fig:training_time}Training time of the LTAI on CORA dataset with stochastic and batch variational inference. Using stochastic variational inference, the per-word predictive log likelihood converges faster than using the batch variational inference.}
\end{figure}

\begin{table}[t!]
\resizebox{ \linewidth}{!}{%

\centering

\begin{tabular}{@{}    l  r  r  r  r  r    @{}}
\hline
Dataset       & \# Tokens & \# Documents & \# Authors & Avg C/D &  Avg C/A \\ \hline
CORA & 17,059 & 13,147& 12,111& 3.46& 12.17\\
Arxiv-Physics & 49,807 & 27,770& 10,950& 12.70& 67.93\\
PNAS & 39,664 & 31,054 & 9,862 & 1.57 & 13.18 \\
Citeseer & 21,223 & 4,255 & 6,384 & 1.24 & 4.38 \\ \hline
\end{tabular}}
\caption{\label{tbl:stats} Datasets. From left to right, each column shows the number of word tokens, number of documents, number of authors, average citations per document (Avg C/D), and average citations per author (Avg C/A).}
\end{table}

\subsection{Updating $\lambda$}
In traditional coordinate ascent based variational inference, the global variational parameter $\lambda$ is updated infrequently because all the other local parameters $\phi$ need to be updated beforehand. This problem is more noticeable in the LTAI since updating $\phi$ using equation \ref{eqn:phi0} is slower than updating $\phi$ in vanilla LDA; moreover, per-author topical authority variable $\eta$ is another local variable that algorithm needs to update a priori. However, using the stochastic variational inference, the global parameters are updated after a small portion of local parameters are updated \cite{hoffman2013stochastic}. Applying stochastic variational inference for the LTAI is straightforward after we calculate the intermediate topic-word variational parameter $\hat{\lambda}$ by $\alpha_\beta + \frac{D}{S_S}\sum^{\mathcal{S}_S}_{d=1}{  \sum^{N_d}_{n=1}{\phi^{k}_{dn} w_{dn}  } }$ from the noisy estimate of the natural gradient with respect to subsampled local parameters where $N_d$ is the number of words for document $d$, and $S_S$ is the subsample size for the minibatch stochastic variational inference. The final global parameter for the ${t}^{th}$ iteration $\lambda^{(t)}$ is updated by $(1-\rho^t)\lambda^{(t-1)} + \rho_t \hat{\lambda}$ where $\rho_t$ is the learning-rate. Posterior inference is guaranteed to converge at local optimum when the learning rate satisfies the condition $\sum^{\infty}_{t=1}{\rho_t}=\infty, \sum^{\infty}_{t=1}{\rho^2_t}<\infty$ \cite{robbins1951stochastic}. In Figure \ref{fig:training_time}, we confirm that stochastic variational inference is applicable for the LTAI and reduces the training time compared to using the batch counterpart, while maintaining similar performance.


\section{Experimental Settings}
\label{sec:exp}

\renewcommand{\arraystretch}{1.3}

In this section, we introduce the four academic corpora used to fit the LTAI, describe comparison models, and provide information about the evaluation metric and parameter settings for the LTAI\footnote{Code and datasets are available at \url{http://uilab.kaist.ac.kr/research/TACL2017/}}.

\subsection{Datasets}
We experiment with four academic corpora: CORA \cite{mccallum2000automating},  Arxiv-Physics \cite{gehrke2003overview}, the Proceedings of the National Academy of Sciences (PNAS), and Citeseer \cite{lu2003link}. 
CORA, Arxiv-Physics, and PNAS datasets contain abstracts only, and the locations of the citations within each paper are not preserved, whereas the Citeseer dataset contains the citation locations. For CORA, Arxiv-Physics, and PNAS, we lemmatize words, remove stop words, and discard words that occur fewer than four times in the corpus. 
Table~\ref{tbl:stats} describes the datasets in detail. Note that we obtain citation data from the entire document, not only from the abstract. Also, we consider within-corpus citation only, which leads to less than 13 average citation counts per document for all corpora. 

\begin{figure}[t!]
    \centering
    \begin{subfigure}[b]{0.49\linewidth}
        \centering
        \includegraphics[width=\linewidth]{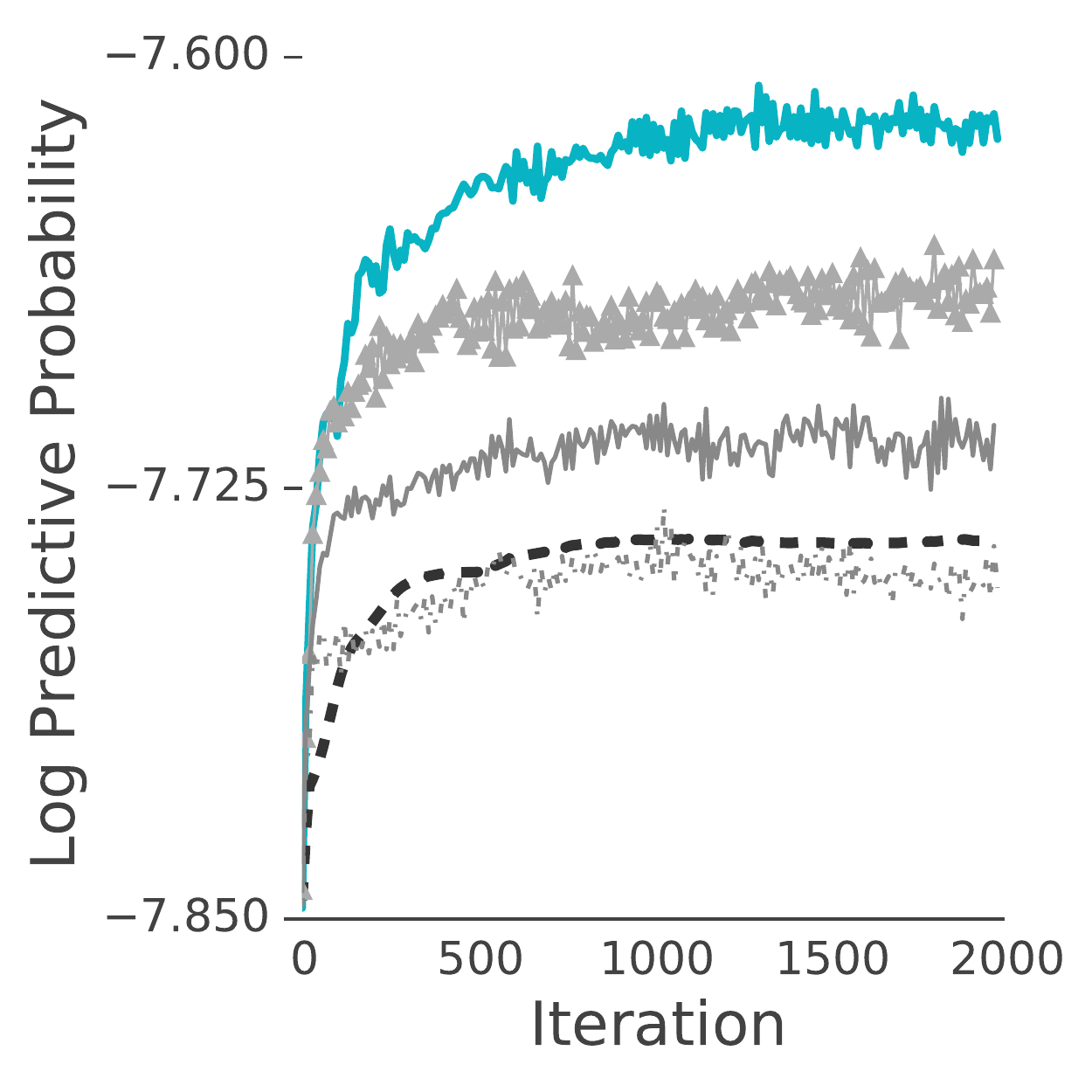}
        \caption{CORA}
        \label{fig:log_predictive_cora}
    \end{subfigure}
    \hfill
    \begin{subfigure}[b]{0.49\linewidth}  
        \centering 
        \includegraphics[width=\linewidth]{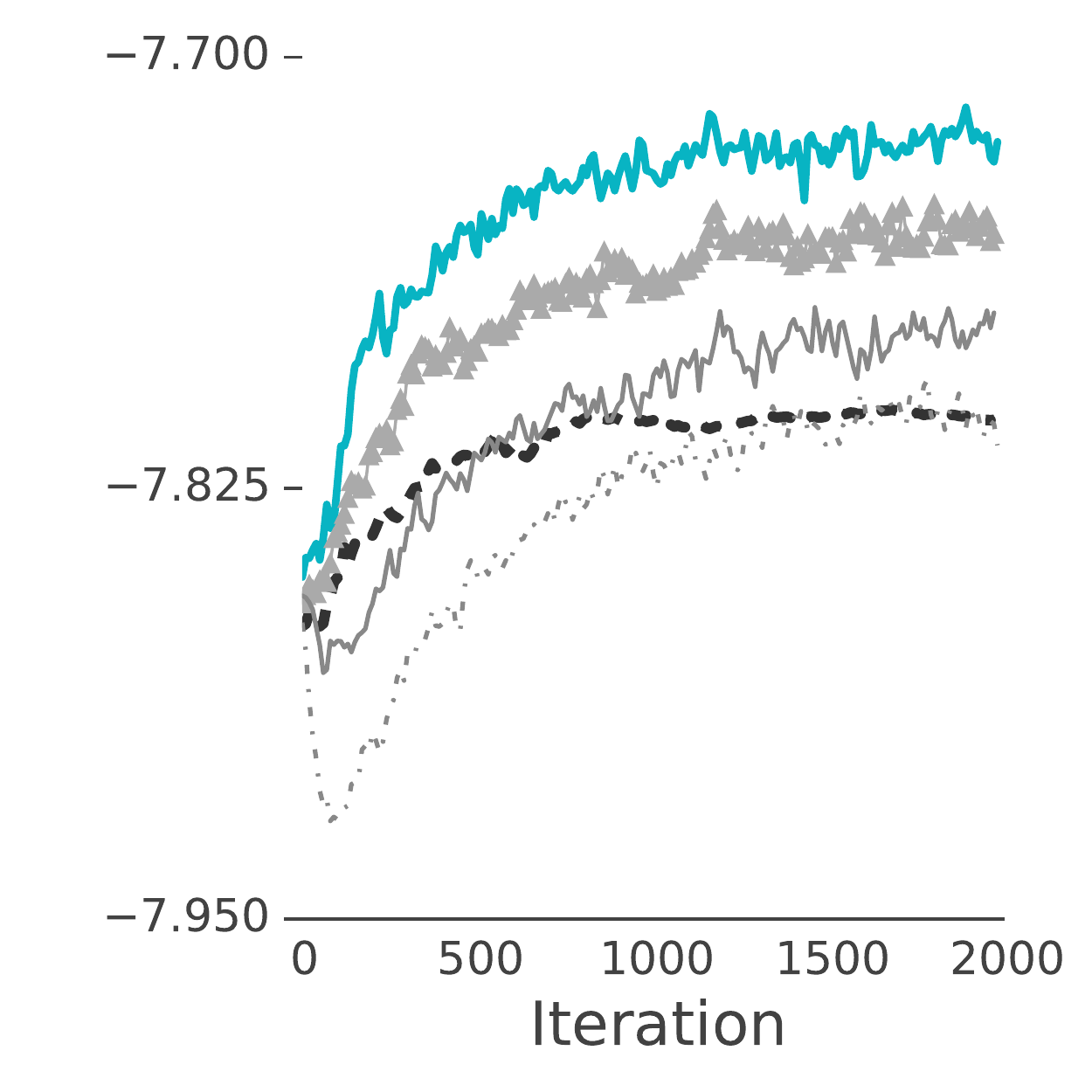}
        \caption{Arxiv-Physics} 
        \label{fig:log_predictive_Arxiv-Physics}
    \end{subfigure}
    \vskip\baselineskip
    
    \begin{subfigure}[b]{0.90\linewidth}   
        \centering 
        \includegraphics[width=\linewidth]{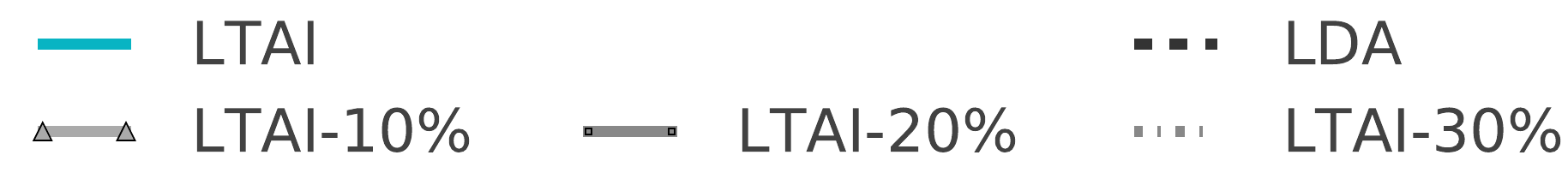}
        
    \end{subfigure}

    \begin{subfigure}[b]{0.49\linewidth}   
        \centering 
        \includegraphics[width=\linewidth]{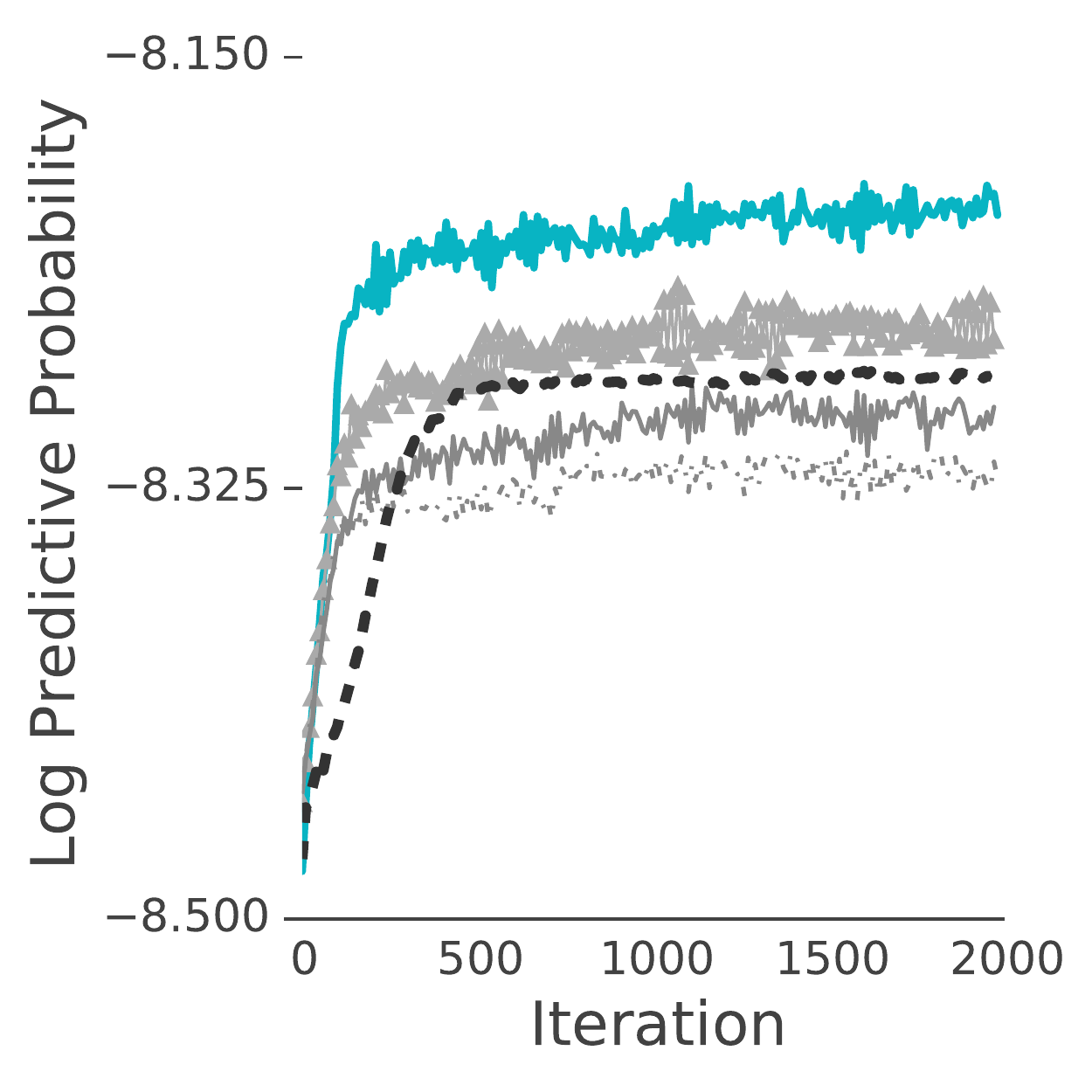}
        \caption{PNAS}
        \label{fig:log_predictive_PNAS}
    \end{subfigure}
    \hfill
    \begin{subfigure}[b]{0.49\linewidth}   
        \centering 
        \includegraphics[width=\linewidth]{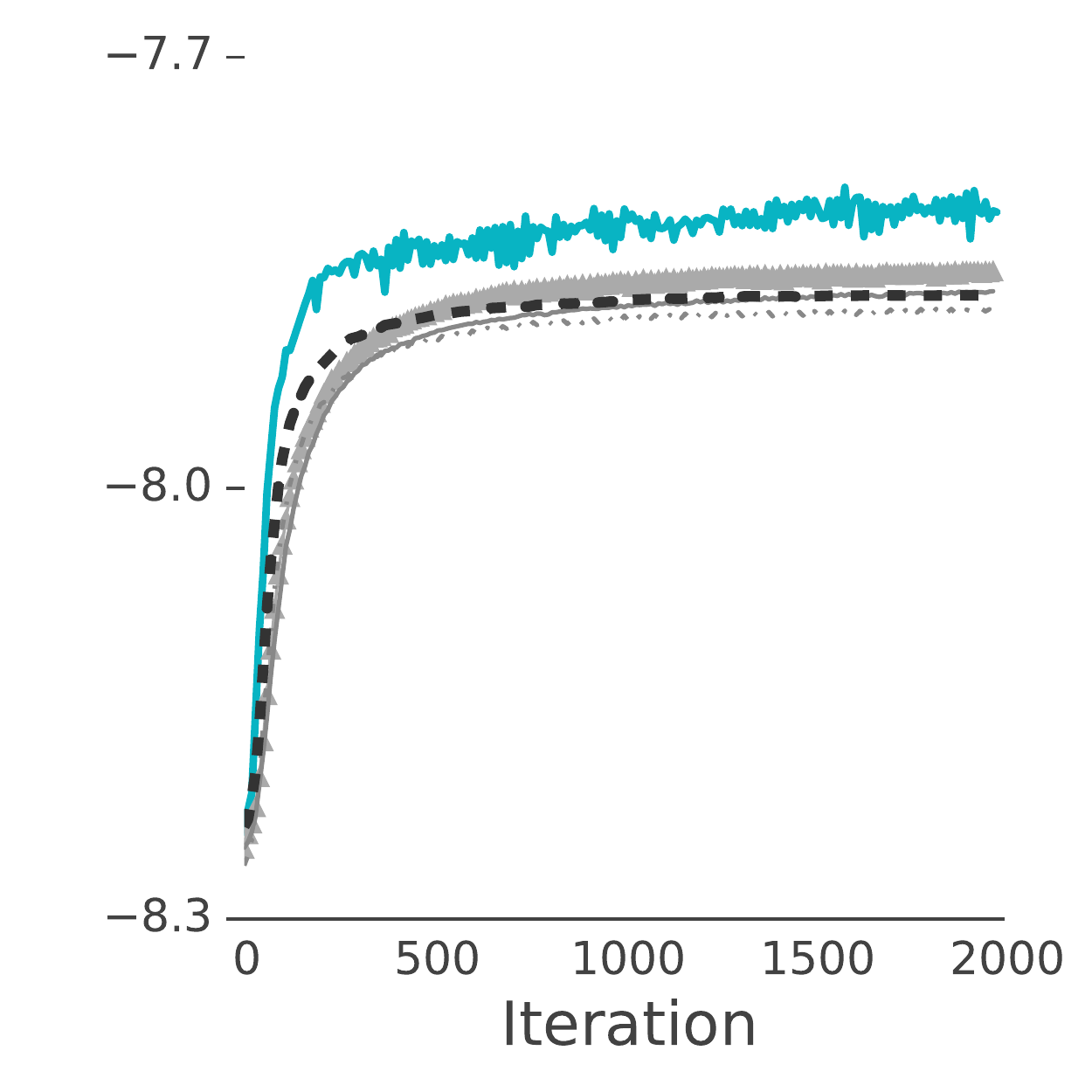}
        \caption{Citeseer}    
        \label{fig:log_predictive_Citeseer}
    \end{subfigure}
    \caption{\label{fig:log_predictive_prediction} Word-level prediction result. We measured per-word log predictive probability on four datasets. As shown in graphs, our model performs better than LDA.}
\end{figure}

\subsection{Comparison Models} 
\label{subsec:comparison_models}
We compare predictive performance of the LTAI with five other models. Different comparison models have different degrees of expressive powers; each model conducts a certain type of prediction task; while RTM, ALTM, and DACTM predicts citation structures, the topical h-index predicts authorship information. Also, the baseline topic models are implemented based on the inference methods suggested in the corresponding papers; LDA, RTM and the LTAI variants use variational inference, while ALTM and DACTM use collapsed Gibbs sampling. Finally, all the conditions for implementation such as the choice of programming language and modules, except for parts that convey each model's unique assumption, are identically set; thus, the performance differences between models are due to their model assumption and different degrees of data usage, rather than the implementation technicalities.

\textbf{Latent Dirichlet Allocation}: LDA \cite{Blei:2003p4796} discovers topics and represents each publication by mixture of the topics. Compared to other models, LDA only uses the content information.

\textbf{LTAI-$n$\%}:  In LTAI-$n$\%, we remove $n$\% of actual citations and displace them with arbitrarily selected false connections. Note that the link structures are displaced rather than removed; if the citation links are just removed, the LTAI and LTAI-$n$\% cannot be fairly compared as the density of the citation structures will be affected and each model needs different concentration values. Performance difference between the LTAI and this indicates that under identical conditions, using the correct linkage information is indeed beneficial for prediction.

\textbf{LTAI-C}: In LTAI-C the precision parameter $c_{ij}$ has constant value, rather than assigning different values according to $x_{ij}$ as discussed in section \ref{sec:model}.

\textbf{LTAI-SEP}: LTAI-SEP has an identical structure as the LTAI, but the topic and the authority variables are separately learned. Once the topic variables are learned using the vanilla LDA, authority and citation variables are then inferred consecutively. Thus, the performance edge of the LTAI over LTAI-SEP highlights the necessity of the LTAI's joint modeling in which both topic and authority related variables reshape one another in an iterative fashion. 

\textbf{Relational Topic Model}: RTM \cite{chang2010hierarchical} jointly models content and citation, and thus, topic proportions of a pair of publications become similar if the pair is connected by citations. Compared to the LTAI, the author information is not considered, the link structure does not have directionality and the model does not consider negative links. 

\textbf{Author-Link Topic Model}: ALTM \cite{kataria2011context} is a variation of author topic model (ATM) \cite{rosen2004author} that models both topical interests and influence of authors in scientific corpora. The model uses content information of citing papers and names of the cited authors as word tokens. ALTM outputs per-topic author distribution that functions as author influence indices.

\textbf{Dynamic Author-Citation Topic Model}: DACTM \cite{kataria2011context} is an extension of ALTM that requires publication corpora which preserve sentence structures. To model author influence, DACTM selectively uses words that are close to the point where the citation is presented. In our corpora, only Citeseer dataset preserves the sentence structure.


\textbf{Topical h-index}: To compute topical h-index, we separate the papers into several clusters using LDA and calculate the h-index within each cluster. Topical h-index is used for author prediction in the same manner as we did for our model, except the topic proportions are replaced to the LDA's result and $\eta$ is replaced to the topical h-index values.

\begin{figure}[t!]
    \centering
    \begin{subfigure}[b]{0.49\linewidth}
        \centering
        \includegraphics[width=\linewidth]{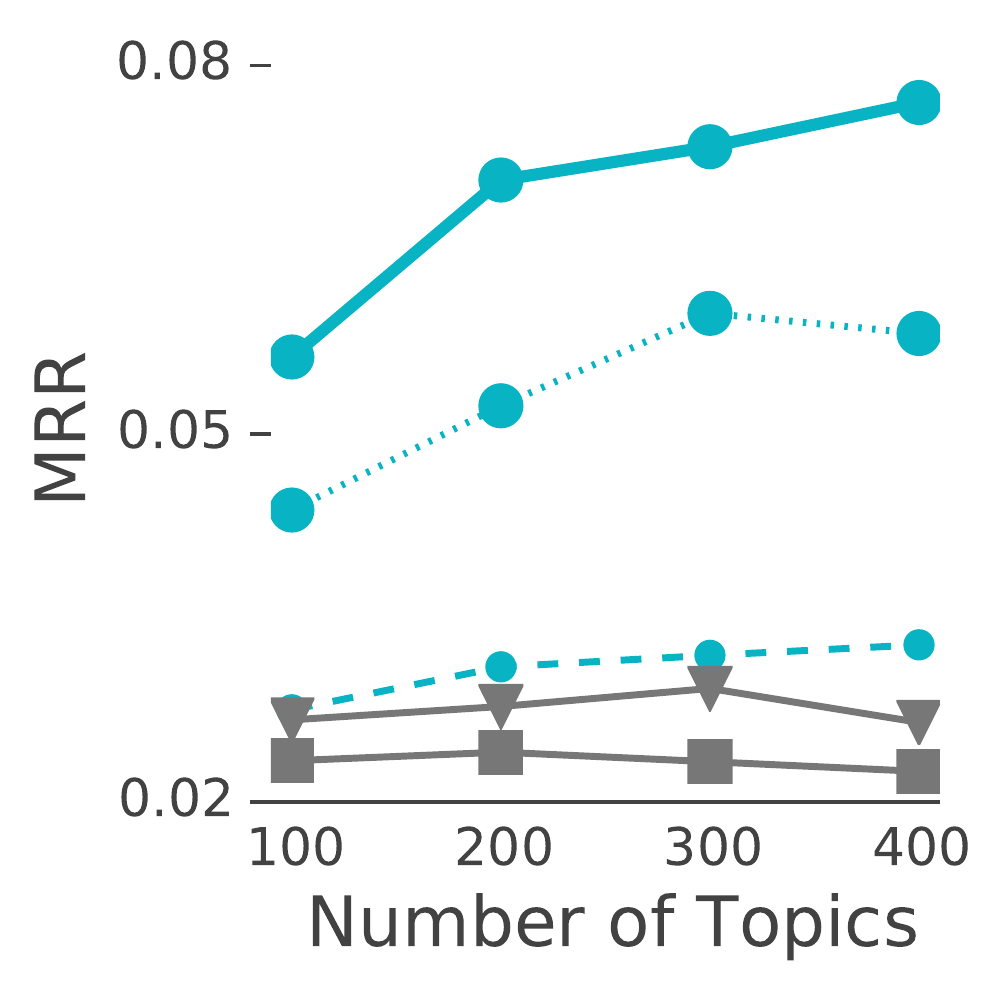}
        \caption{CORA}
        \label{fig:link_cora}
    \end{subfigure}
    \hfill
    \begin{subfigure}[b]{0.49\linewidth}  
        \centering 
        \includegraphics[width=\linewidth]{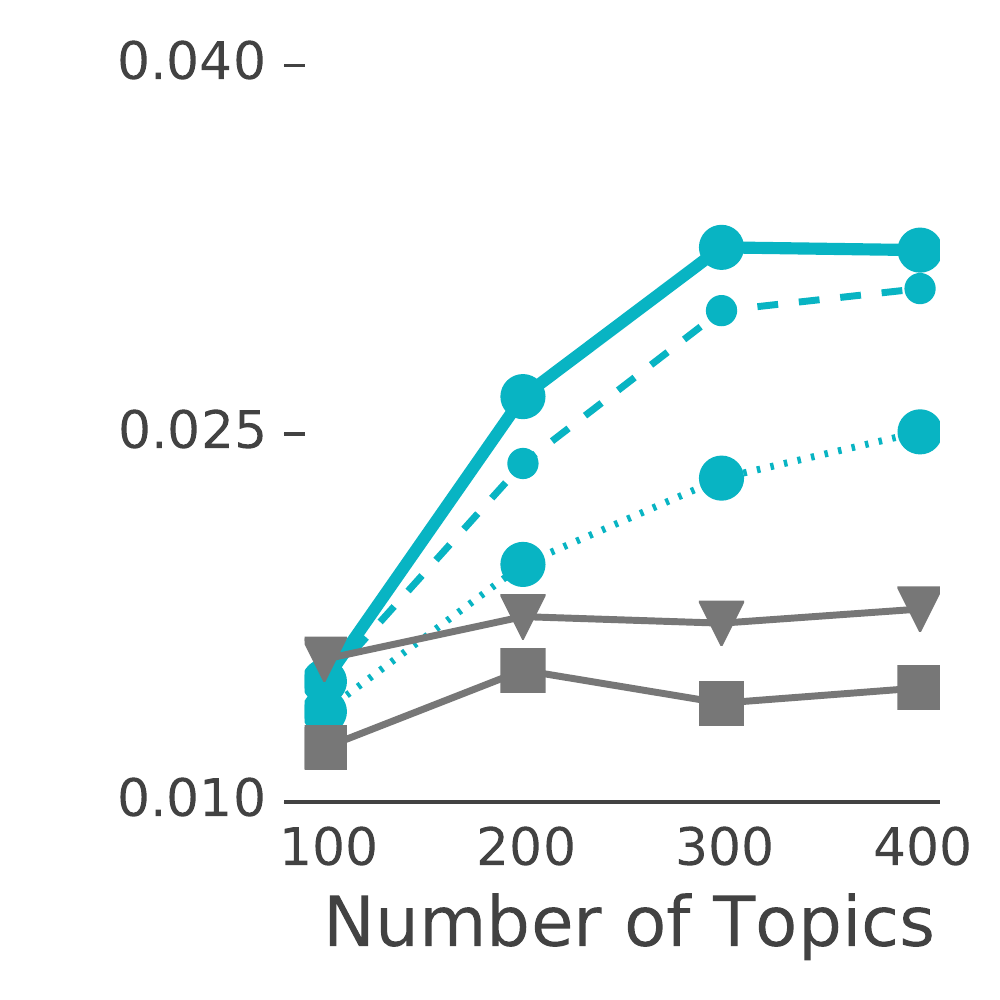}
        \caption{Arxiv-Physics} 
        \label{fig:link_Arxiv-Physics}
    \end{subfigure}
    \vskip\baselineskip
    
    \begin{subfigure}[b]{0.7\linewidth}   
        \centering 
        \includegraphics[width=\linewidth]{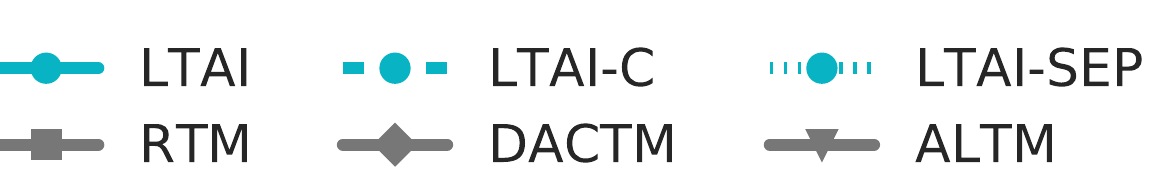}
        
    \end{subfigure}

    \vskip\baselineskip

    \begin{subfigure}[b]{0.49\linewidth}   
        \centering
        \includegraphics[width=\linewidth]{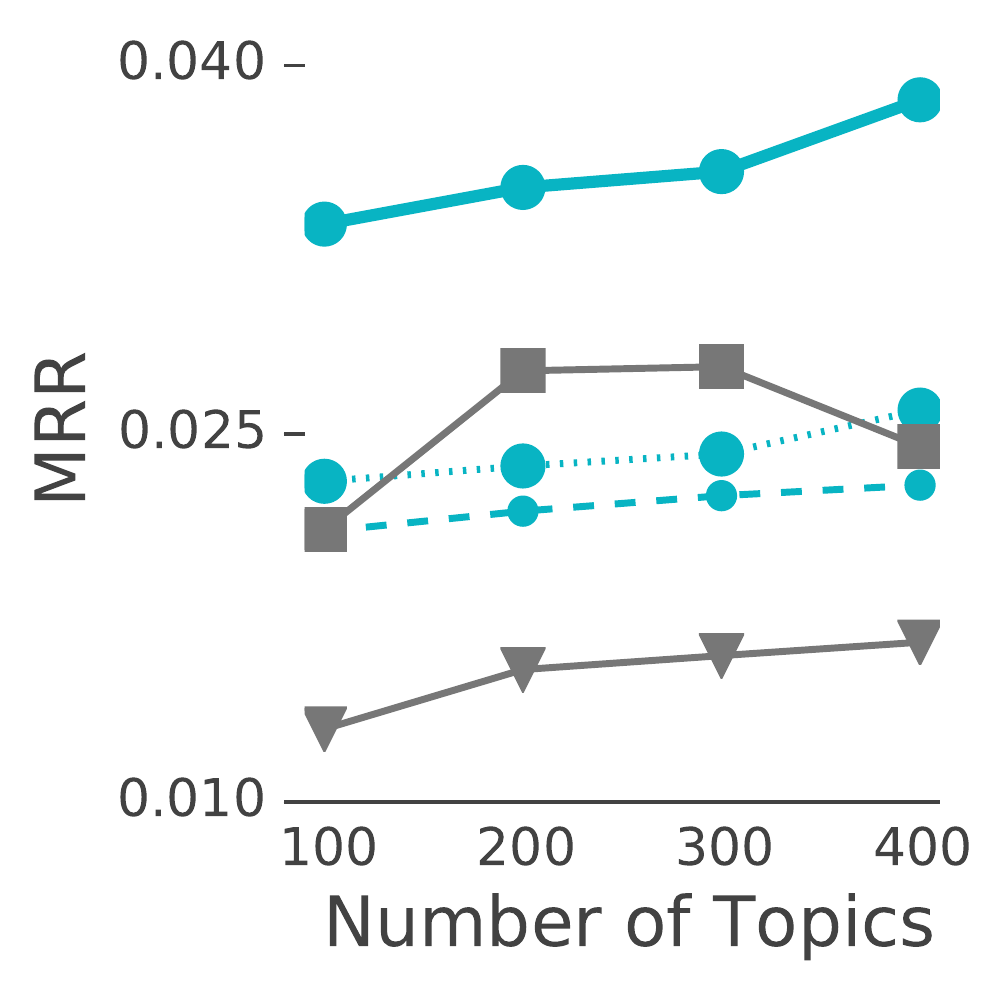}
        \caption{PNAS}
        \label{fig:link_PNAS}
    \end{subfigure}
    \hfill
    \begin{subfigure}[b]{0.49\linewidth}   
        \centering 
        \includegraphics[width=\linewidth]{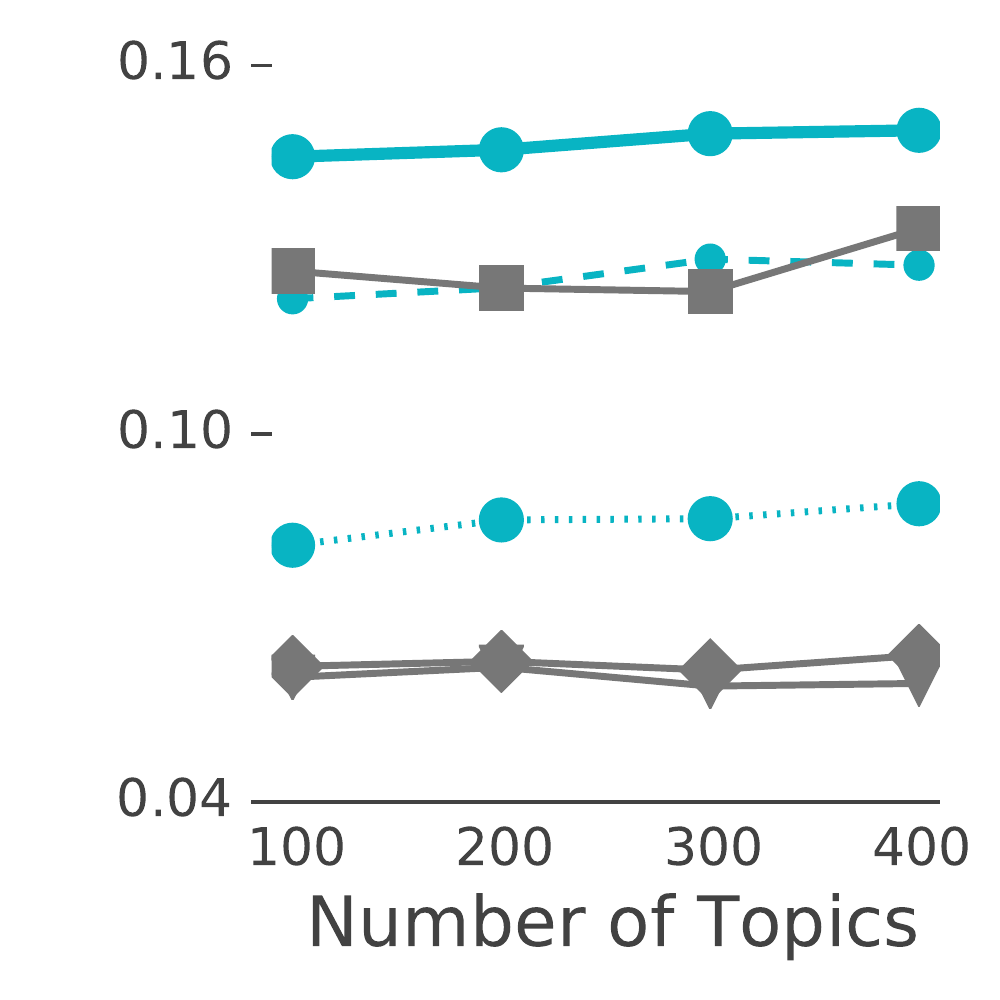}
        \caption{Citeseer}    
        \label{fig:link_Citeseer}
    \end{subfigure}
    \caption{\label{fig:link_prediction} Citation prediction results. The task is to find out which paper is originally linked to a cited paper. We measure mean reciprocal rank (MRR) to evaluate model performance. For all cases, the LTAI performs better than the other methods.}
\end{figure}

\subsection{Evaluation Metric and Parameter Settings}
We use mean reciprocal rank (MRR) \cite{voorhees1999trec} to measure the predictive performance of the LTAI and the comparison models. MRR is a widely used metric for evaluating link prediction tasks \cite{balog2007determining,diehl2007relationship,radlinski2008does,huang2015neural}. When the models outputs the correct answers as ranks, MRR is the inverse of the harmonic mean of such ranks.

We report the parameter values used for evaluations. For all datasets, we set $c_-$ to 1. To predict citation, we set $c_+$ to 10,000, 100, 1,000, 10, and to predict authorship, we set $c_+$ to 1,000, 1,000, 10,000, 1,000 for CORA, Arxiv-Physics, PNAS, and Citeseer datasets. These values are obtained through exhaustive parameter analysis. We set $\alpha_{\theta}$ to $1$, and $\alpha_{\beta}$ to $0.1$. We fix the subsample sizes to 500\footnote{Although we do not present thorough sensitivity analysis in this paper, we confirm that the performance of our model was robust against adjusting the parameters within a factor of 2.}. For fair comparison, all the parameters that the LTAI and the baseline models share are set to have the same values, and for other parameters that uniquely belong to the baseline models, the values are exhaustively tuned as done in the LTAI. Finally, we note that all parameters are tuned using the training set, and test dataset is used only for the testing purpose.


\section{Evaluation}
\label{sec:eval}

We conduct the evaluation of the LTAI with three different quantitative tasks, along with one qualitative analysis. In the first task, we check whether using citation and authorship information in the LTAI helps increase the word-level predictive performance. In the second and third tasks, we measure the predictability of the LTAI regarding missing publication-publication linkage and author-publication linkage; with these two tasks, we compare the predictive power of the LTAI with other comparison models and use MRR as evaluation metric. Finally, we observe famous researchers' topical authority scores generated by the LTAI and investigate how these scores capture notable academic characteristics of the researchers.

\begin{figure}[t!]
    \centering
    \begin{subfigure}[b]{0.49\linewidth}
        \centering
        \includegraphics[width=\linewidth]{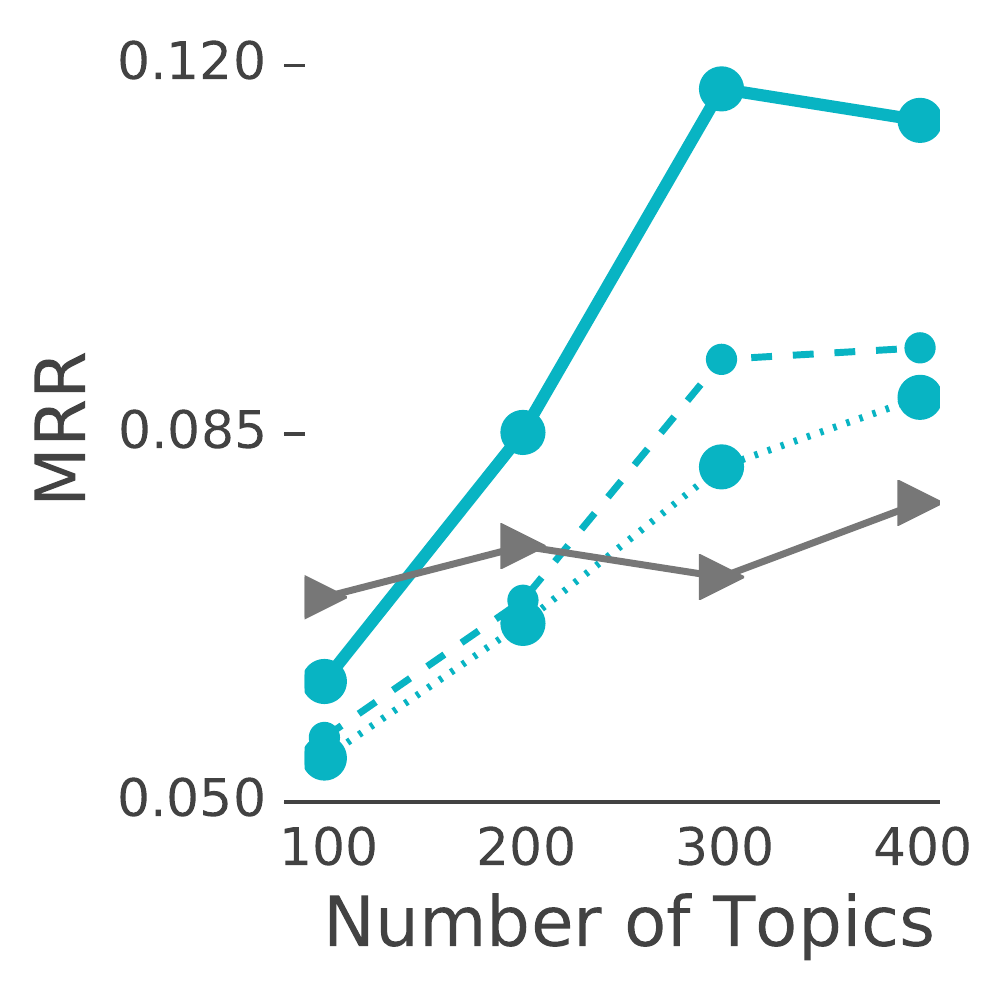}
        \caption{CORA}
        \label{fig:author_cora}
    \end{subfigure}
    \hfill
    \begin{subfigure}[b]{0.49\linewidth}  
        \centering 
        \includegraphics[width=\linewidth]{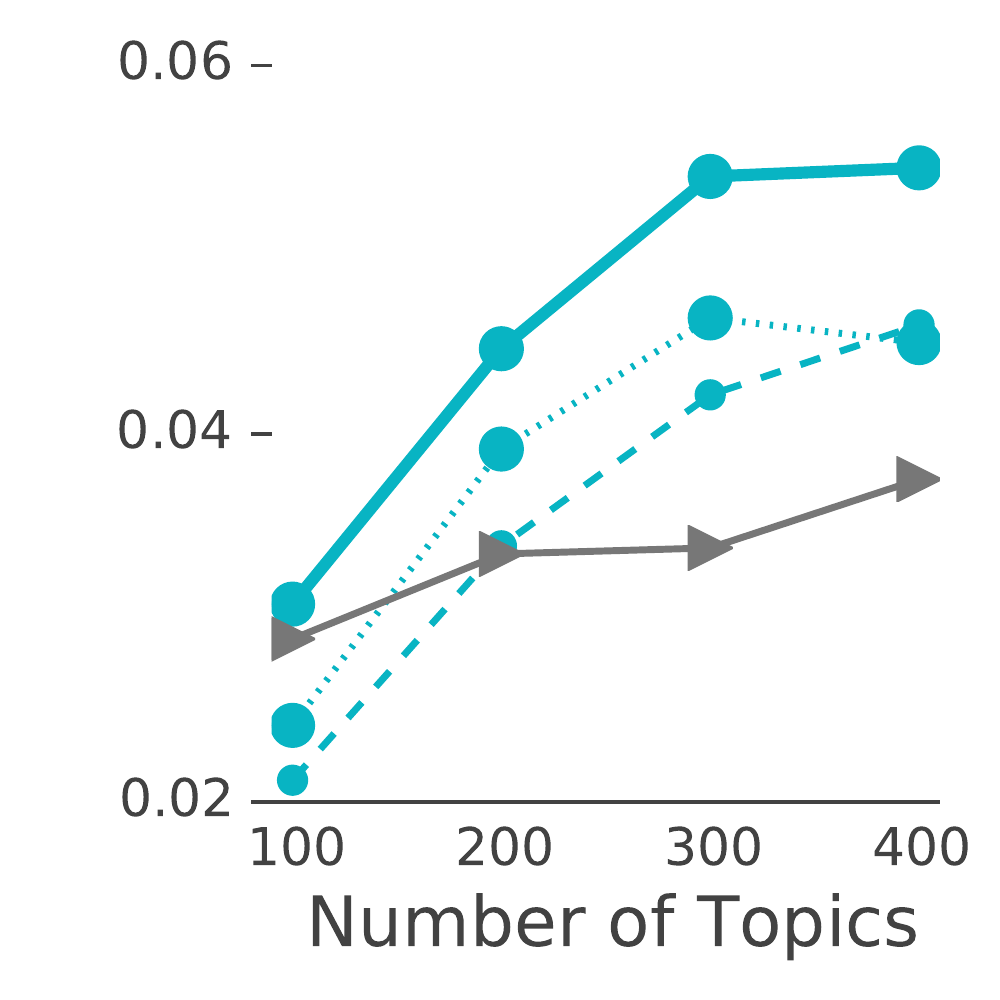}
        \caption{Arxiv-Physics} 
        \label{fig:author_Arxiv-Physics}
    \end{subfigure}
    \vskip\baselineskip
    
    \begin{subfigure}[b]{\linewidth}   
        \centering 
        \includegraphics[width=0.9\linewidth]{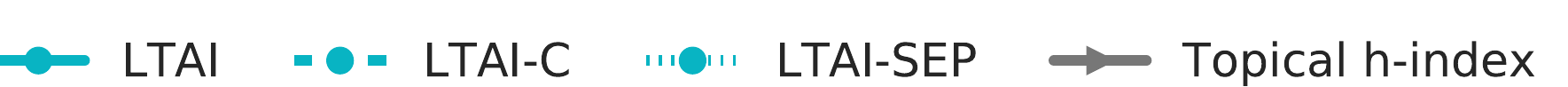}
        
    \end{subfigure}

    \vskip\baselineskip
    
    \begin{subfigure}[b]{0.49\linewidth}   
        \centering 
        \includegraphics[width=\linewidth]{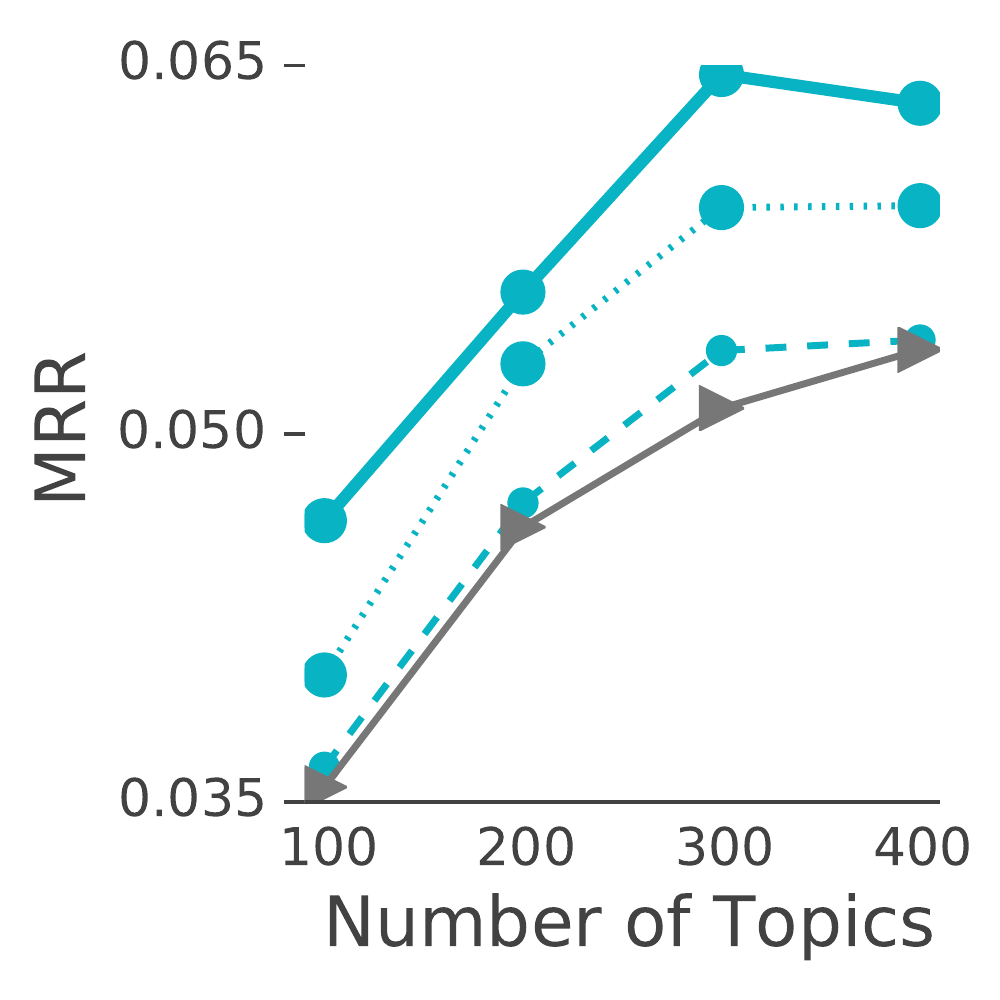}
        \caption{PNAS}
        \label{fig:author_PNAS}
    \end{subfigure}
    \hfill
    \begin{subfigure}[b]{0.49\linewidth}   
        \centering 
        \includegraphics[width=\linewidth]{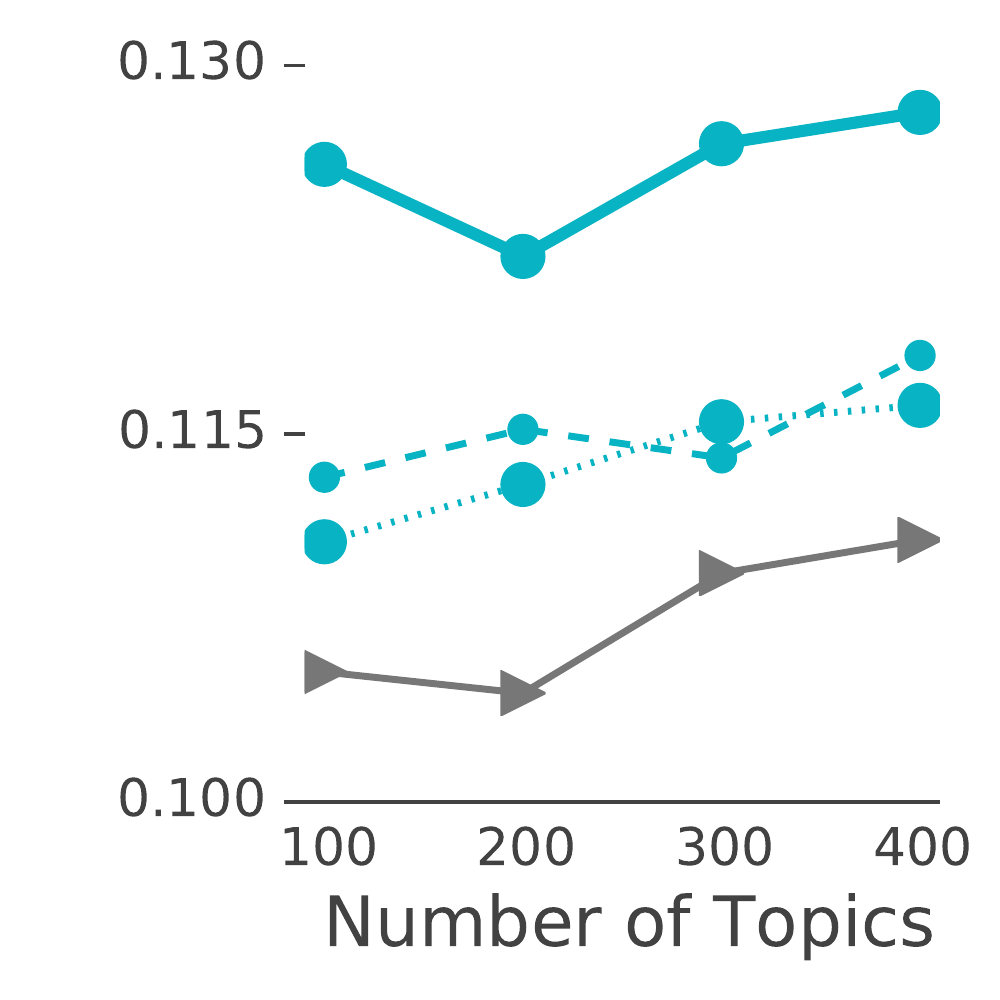}
        \caption{Citeseer}    
        \label{fig:author_Citeseer}
    \end{subfigure}
    \caption{\label{fig:author_prediction} Author prediction results. The task is to find out who the author of a cited paper is, given all the citing papers. For all cases, the LTAI performs better than the other methods.}
\end{figure}

\subsection{Word-level Prediction}
In the LTAI, citation and authorship information affect per-document topic proportions, as can be confirmed in equation \ref{eqn:phi0}. This joint modeling of content and linkage structure, compared to vanilla LDA that uses content data only, yields better performance in terms of predicting missing words in documents. In this task, we use log-predictive probability, a metric that is widely used in other researches for measuring model fitness \cite{Teh:2006p3792,asuncion2009smoothing,hoffman2013stochastic}. For each corpus, we separate one third of documents as test set, and for all documents in each test set, we use half of the words for training per-document topic proportion $\theta$ and predict the probability of word occurrence regarding the remaining half. Specifically, the predictive probability for a word in a test set $w_{new}$ with respect to the given words $w_{obs}$ and the training document $\mathcal{D}_{train}$ is computed using equation $p(w_{new} | \mathcal{D}_{train}, w_{obs}) = \sum^{K}_{k=1}{\mathbb{E}_q[\theta_k] \mathbb{E}_q[\beta_{k, w_{new}}]}$.

Figure \ref{fig:log_predictive_prediction} illustrates the per-word log-predictive probability in each corpus. We confirm that when using the LTAI, the log predictive probability converges at higher value compared to the result using LDA. Also, when we corrupt the link structure from $10\%$ to $30\%$ the predictive performances of the LTAI gradually decrease. Thus, the LTAI's superior predictive performance is attributed to its usage of correct citations rather than the algorithmic bias.

\subsection{Citation Prediction}

We evaluate model predictability regarding which publication is originally citing a certain publication. Specifically, we randomly remove one citation from each of the documents in the test set. To predict the citation link between publications, we first compute the probability that publication $j$ cites $i$ from $p(x_{i \leftarrow j}| z, \mathcal{A}_i, \pi_i) \propto \sum_{a\in \mathcal{A}_i} \pi_{i \leftarrow j a} \mathcal{N}(x_{i \leftarrow j}|\bar{z}_i^{\top} \text{diag}(\eta_a) \bar{z}_j, c_+^{-1})$. Given the topic proportion of the cited publication $\theta_i$ and the topical authorities of the authors $\eta_a$, we compute which publication is more likely to cite the publication. Based on our model assumption in subsection \ref{subsec:citation_generation}, using topical authority increases the performance of predicting linkage structure. 

In Figure \ref{fig:link_prediction}, the LTAI yields better citation prediction performance than other models for all datasets and with most number of topics. Since the LTAI incorporates topical authority for predicting citations, it performs better than RTM, which does not discover topical authority. We can attribute the better performance of the LTAI compared to ALTM and DACTM to the LTAI's multiple model assumptions explained in section \ref{sec:model}. We note that DACTM requires additional information such as citation location and sentence structure, and thus, is only applicable for limited kinds of datasets. 

\begin{table*}
\centering
\resizebox{0.7 \linewidth}{!}{%
\begin{tabular}{@{} l r r r l r@{}}
\toprule
           
Author   & h-index & \# cite &\# paper  & Representative Topic  & T Authority \\ \hline

D Padua & 12                                                                           & 291                           & 21                         & parallel, efficient, computation, runtime  & 10.36 \\
V Lesser & 11                                                                           & 303                           & 48                         & interaction, intelligent, multiagent, autonomous & 11.92  \\
M Lam & 11                                                                           & 440                           & 20                         & memory, processor, cache, synchronization & 12.74\\
M Bellare & 11                                                                           & 280                           & 43                         & scheme, security, signature, attack  & 13.21\\
L Peterson & 10                                                                           & 297                           & 24                         & operating, mechanism, interface, thread  & 9.28\\
D Ferrari & 10                                                                           & 377                           & 18                         & traffic, delay, bandwidth, allocation  & 14.16\\
O Goldreich & 9                                                                           & 229                           & 49                         & proof, known, extended, notion & 12.57 \\
M Jordan & 9                                                                           & 263                           & 27                         & approximation, intelligence, artificial, correlation  & 10.15\\
D Culler & 9                                                                           & 565                           & 30                         & operating, mechanism, interface, thread  & 12.37 \\
A Pentland  & 8                                                                           & 207                           & 39                         & image, motion, visual, estimate & 10.82\\ \bottomrule
\end{tabular}
}
\caption{\label{tbl:qan_cora_overall} Authors with highest h-index scores and their statistics from the CORA dataset. We show the authors with their  h-index, number of citations (\# cite), and number of papers (\# paper), representative topic, and their topical authority (T Authority) of the corresponding topic. We show that while the authors have the highest h-indices with lots of papers written and lots of citations earned, the topics which the authors exert authority varies.}
\end{table*}

\subsection{Author Prediction}

For author prediction, we randomly remove one of the authors from documents in the test set while preserving citation structures. Similar to citation prediction, we predict which author is more likely to write the cited publication based on the topic proportions of cited publication $i$ and a set of citing publications $\mathcal{J}$. We approximate the probability of researcher $a$ being an author of publication $i$ from $p(a | z, \eta_a, x_{i \leftarrow j}) \propto \prod_{j\in \mathcal{J}}  \mathcal{N}(x_{i \leftarrow j}|\bar{z}_i^{\top} \text{diag}(\eta_a) \bar{z}_j, c_+^{-1})$. Because the mixture proportion of an unknown author $\pi_{i \leftarrow j a}$ cannot be obtained during posterior inference, we assume the cited publication is written by a single author to approximate the probability. For author prediction, we choose the author that maximizes the above probability. In Figure \ref{fig:author_prediction}, the LTAI outperforms the comparison models in most of the settings.

\subsection{Qualitative Analysis}

To stress our model's additional characteristics that are not observed in the quantitative analysis, we look at the assigned topical authority indices as well as other statistics of some researchers in the dataset. In the analyses, we set the number of topics to 100, and use CORA dataset for demonstration.

We first demonstrate famous authors' authoritative topics that can be unveiled using our model. In \autoref{tbl:qan_cora_overall}, we list top 10 authors with highest h-indices along with their number of citations, number of papers, and their representative topics. Authors' representative topics are the topics with highest authority scores. In the table, we observe that all authors with top h-indices have wrote at least 18 papers and earned at least 207 citations, which are the top 0.8\% and 0.2\% values respectively. However, their authoritative topics retrieved by the LTAI do not overlap for any of the authors. This table illustrates that each of the top authors in the table exerts authority on different academic topics that can be captured by the LTAI, while the authors commonly have highest h-index scores as well as other statistics.

 We now stress attributes of topical authority index that are different from other topic irrelevant statistics. From Tables \ref{tbl:qan_cora_cv} to \ref{tbl:qan_cora_cs}, we show four example topics extracted by our model and list notable authors within each topic with their topical authority indices, h-indices, number of citations, and number of papers. In the tables, we first find that all four authors with highest topical authority values, Monica Lam, Alex Pentland, Michael Jordan, and Mihir Bellare are also listed in the topic-irrelevant authority rankings in \autoref{tbl:qan_cora_overall}. From this, we confirm that authority score of the LTAI has a certain degree of correlation to other statistics, while it splits the authors by their authoritative topics.
 
 At the same time, the topical authority score correlates less with topic-irrelevant statistics than those statistics correlate with themselves; in \autoref{tbl:qan_cora_cs}, Oded Goldreich has lower topical authority score for the computer security topic while having higher topic irrelevant scores than the above four researchers, because his main research filed is in the theory of computation and randomness. Also, we can spot authors who exert high authority on multiple academic fields, such as Tomaso Poggio in \autoref{tbl:qan_cora_cv} and in \autoref{tbl:qan_cora_ai}. Similarity, when comparing Federico Girosi and Tomaso Poggio in \autoref{tbl:qan_cora_ai}, the two researchers have similar authority indices for this topic while Tomaso Poggio has higher values for the other three topic-irrelevant indices. This is a reasonable outcome when we investigate the two researchers' publication history. Federico Girosi has relatively focused academic interest, with his publication history being skewed towards machine-learning-related subjects, while Tomaso Poggio has broader topical interests that include computer vision and statistical learning, while also co-authoring most of the papers that Federico Girosi wrote. Thus, Federico Girosi has similar authority index for this topic but has lower authority indices for other topics than Tomaso Poggio.
 
Also, our model is able to capture topic-specific authoritative researchers that have relatively low topic-irrelevant scores. For example, researchers such as Stan Sclaroff and Kentaro Toyama are the top 5 authoritative researchers in computer vision topic according to the LTAI, but it is difficult to detect these researchers out of many other authoritative authors using the topic-irrelevant scores.

Finally, the LTAI detect researchers' topical authority that is peripheral but not negligible. Mark Jones in \autoref{tbl:qan_cora_ai}, who has high h-index, number of citations, and wrote many papers, is a researcher whose academic interest lies in programming language design and application. However, while most of his papers' main topics are about programming language, he often uses inference techniques and algorithms in machine learning in his papers. Our model captures that tendency and assigns some authority score for machine learning to him.

\section{Conclusion and Discussion}
\label{sec:discussions}




\begin{table}[t!]
\centering
\resizebox{0.44 \textwidth}{!}{%
\begin{tabular}{@{} r l r r r r @{}}
\toprule
\multicolumn{6}{c}{\begin{tabular}[c]{@{}c@{}}Topic: image, motion, visual, estimate, \\ robust, shape, scene, geometric \end{tabular}}           \\ \hline \hline

Rank & Author   & Topical Authority & h-index & \# cite &\# paper \\ \hline

1 & A Pentland & 10.82                                                                           & 8                           & 207                         & 39                           \\
2 &J Fessler & 9.09                                                                           & 6                           & 92                         & 26                           \\
3 & T Poggio & 8.22                                                                           & 6                           & 178                         & 27                           \\
4 & S Sclaroff & 7.61 & 3 & 69 & 11 \\
5 & K Toyama  & 6.65                                                                           & 4                           & 41                         & 10            \\ \bottomrule
\end{tabular}
}
\caption{\label{tbl:qan_cora_cv} Authors who have high authority score in computer architecture topic computer vision topic.}
\end{table}

\begin{table}[t!]
\centering
\resizebox{0.445 \textwidth}{!}{%
\begin{tabular}{@{}r l r r r r @{}}
\toprule
\multicolumn{6}{c}{\begin{tabular}[c]{@{}c@{}}Topic: approximation, intelligence, artificial, \\ correlation, support, recognition, model, representation\end{tabular}}           \\ \hline \hline

Rank & Author   & Topical Authority & h-index & \# cite &\# paper \\ \hline

1 & M Jordan & 10.15                                                                           & 9                           & 263                         & 27                           \\
2 & M Warmuth & 9.57                                                                           & 8                           & 160                         & 17                           \\
13 & T Poggio & 3.48                                                                           & 6                           & 178                         & 27                           \\
17 & F Girosi & 3.22                                                                           & 3                           & 101                         & 9                           \\
34 & M Jones  & 2.06                                                                           & 7                           & 151                         & 20            \\ \bottomrule
\end{tabular}
}
\caption{\label{tbl:qan_cora_ai} Authors who have high authority score in artificial intelligence topic.}
\end{table}

\begin{table}[t!]
\centering
\resizebox{0.45 \textwidth}{!}{%
\begin{tabular}{@{} r l r r r r @{}}
\toprule
\multicolumn{6}{c}{\begin{tabular}[c]{@{}c@{}}Topic: scheme, security, signature, attack, \\ threshold, authentication, cryptographic, encryption \end{tabular}}           \\ \hline \hline

Rank & Author   & Topical Authority & h-index & \# cite &\# paper \\ \hline

1 & M Bellare & 13.21                                                                           & 11                           & 280                         & 43                           \\
2 &P Rogaway & 11.98                                                                           & 7                           & 117                         & 13                           \\
3 & H Krawczyk & 7.29                                                                           & 6                           & 75                         & 15                           \\
4 & R Canetti & 7.13 & 4 & 40 & 10 \\
9 & O Goldreich  & 3.70                                                                           & 9                           & 229                         & 49            \\ \bottomrule
\end{tabular}
}
\caption{\label{tbl:qan_cora_cs} Authors who have high authority score in computer security topic.}
\end{table}

We proposed Latent Topical Authority Indexing (LTAI) to model the topical-authority of academic researchers. Based on the hypothesis that authors play an important role in citation, we specifically focus on their authority and develop a Bayesian model to capture the authority. With model assumptions that are necessary for extracting convincing and interpretable topical authority values for authors, we have proposed speed-up methods that are based on stochastic optimization.

While there is prior research in topic modeling that provides topic-specific indices when modeling the link structure, these do not extend to individual indices, and most previous citation-based indices are defined for each individual but without considering topics. On the other hand, our model combines the merits of both topic-specific and individual-specific indices to provide topical authority information for academic researchers.

With four academic datasets, we demonstrated that the joint modeling of publication and author related variables improve topic quality, when compared to vanilla LDA. Also, we quantitatively manifested that including authority variables increases the predictive performance in terms of citation and author predictions. Finally, we qualitatively demonstrated the interpretability by topical-authority outcomes of the LTAI from the CORA corpus. 

Finally, there are issues that can be dealt in future work. In our model, we do not consider time information in terms of when papers are published and when pairs of papers are linked; we can use datasets that incorporate timestamps to enhance the model capability to predict future citations and authorships.

\clearpage

\section*{Acknowledgments}

We thank Jae Won Kim for his help on collecting, refining the dataset and contributing to the early version of the manuscript, anonymous reviewers as well as the TACL editor Noah Smith for detailed and thoughtful comments, and Joon Hee Kim and other UILab members for providing helpful insights in the research. This work was supported by Institute for Information \& communications Technology Promotion(IITP) grant funded by the Korea government(MSIP) (No.B0101-15-0307, Basic Software Research in Human-level Lifelong Machine Learning (Machine Learning Center)).

\bibliographystyle{acl2012}
\bibliography{jykim}

\begin{thebibliography}{}

\bibitem[\protect\citename{Asuncion \bgroup et al.\egroup
  }2009]{asuncion2009smoothing}
Arthur Asuncion, Max Welling, Padhraic Smyth, and Yee~Whye Teh.
\newblock 2009.
\newblock On smoothing and inference for topic models.
\newblock In {\em UAI}.

\bibitem[\protect\citename{Balog and De~Rijke}2007]{balog2007determining}
Krisztian Balog and Maarten De~Rijke.
\newblock 2007.
\newblock Determining expert profiles (with an application to expert finding).
\newblock In {\em IJCAI}.

\bibitem[\protect\citename{Blei \bgroup et al.\egroup }2003]{Blei:2003p4796}
David~M. Blei, Andrew~Y. Ng, and Michael~I. Jordan.
\newblock 2003.
\newblock Latent {D}irichlet allocation.
\newblock {\em JMLR}, pages 993--1022.

\bibitem[\protect\citename{Caragea \bgroup et al.\egroup
  }2014]{caragea2014citation}
Cornelia Caragea, Florin~Adrian Bulgarov, Andreea Godea, and Sujatha~Das
  Gollapalli.
\newblock 2014.
\newblock Citation-enhanced keyphrase extraction from research papers: A
  supervised approach.
\newblock In {\em EMNLP}.

\bibitem[\protect\citename{Caragea \bgroup et al.\egroup }2015]{caragea2015co}
Cornelia Caragea, Florin Bulgarov, and Rada Mihalcea.
\newblock 2015.
\newblock Co-training for topic classification of scholarly data.
\newblock In {\em EMNLP}.

\bibitem[\protect\citename{Chang and Blei}2010]{chang2010hierarchical}
Jonathan Chang and David~M. Blei.
\newblock 2010.
\newblock Hierarchical relational models for document networks.
\newblock {\em The Annals of Applied Statistics}, pages 124--150.

\bibitem[\protect\citename{Cohan and Goharian}2015]{cohan2015scientific}
Arman Cohan and Nazli Goharian.
\newblock 2015.
\newblock Scientific article summarization using citation-context and article's
  discourse structure.
\newblock In {\em EMNLP}.

\bibitem[\protect\citename{Diehl \bgroup et al.\egroup
  }2007]{diehl2007relationship}
Christopher~P. Diehl, Galileo Namata, and Lise Getoor.
\newblock 2007.
\newblock Relationship identification for social network discovery.
\newblock In {\em AAAI}.

\bibitem[\protect\citename{Dietz \bgroup et al.\egroup
  }2007]{Dietz:2007:UPC:1273496.1273526}
Laura Dietz, Steffen Bickel, and Tobias Scheffer.
\newblock 2007.
\newblock Unsupervised prediction of citation influences.
\newblock In {\em ICML}.

\bibitem[\protect\citename{Dong \bgroup et al.\egroup }2015]{dong2015will}
Yuxiao Dong, Reid~A. Johnson, and Nitesh~V. Chawla.
\newblock 2015.
\newblock Will this paper increase your h-index?: Scientific impact prediction.
\newblock In {\em WSDM}.

\bibitem[\protect\citename{Foulds and Smyth}2013]{foulds2013modeling}
James~R. Foulds and Padhraic Smyth.
\newblock 2013.
\newblock Modeling scientific impact with topical influence regression.
\newblock In {\em EMNLP}.

\bibitem[\protect\citename{Garfield}2006]{garfield2006history}
Eugene Garfield.
\newblock 2006.
\newblock The history and meaning of the journal impact factor.
\newblock {\em JAMA}, 295(1):90--93.

\bibitem[\protect\citename{Gehrke \bgroup et al.\egroup
  }2003]{gehrke2003overview}
Johannes Gehrke, Paul Ginsparg, and Jon Kleinberg.
\newblock 2003.
\newblock Overview of the 2003 {KDD} {C}up.
\newblock {\em ACM SIGKDD Explorations Newsletter}, 5(2):149--151.

\bibitem[\protect\citename{He \bgroup et al.\egroup }2015]{he2015discovering}
Yuan He, Cheng Wang, and Changjun Jiang.
\newblock 2015.
\newblock Discovering canonical correlations between topical and topological
  information in document networks.
\newblock In {\em CIKM}.

\bibitem[\protect\citename{Hirsch}2005]{hirsch2005index}
Jorge~E. Hirsch.
\newblock 2005.
\newblock An index to quantify an individual's scientific research output.
\newblock {\em PNAS}, 102(46):16569--16572.

\bibitem[\protect\citename{Hoffman \bgroup et al.\egroup
  }2013]{hoffman2013stochastic}
Matthew~D. Hoffman, David~M. Blei, Chong Wang, and John~W. Paisley.
\newblock 2013.
\newblock Stochastic variational inference.
\newblock {\em JMLR}, 14(1):1303--1347.

\bibitem[\protect\citename{Hu \bgroup et al.\egroup }2008]{hu2008collaborative}
Yifan Hu, Yehuda Koren, and Chris Volinsky.
\newblock 2008.
\newblock Collaborative filtering for implicit feedback datasets.
\newblock In {\em ICDM}.

\bibitem[\protect\citename{Huang \bgroup et al.\egroup }2015]{huang2015neural}
Wenyi Huang, Zhaohui Wu, Chen Liang, Prasenjit Mitra, and C.~Lee Giles.
\newblock 2015.
\newblock A neural probabilistic model for context based citation
  recommendation.
\newblock In {\em AAAI}.

\bibitem[\protect\citename{Jiang}2015]{jiang2015chronological}
Zhuoren Jiang.
\newblock 2015.
\newblock Chronological scientific information recommendation via supervised
  dynamic topic modeling.
\newblock In {\em WSDM}.

\bibitem[\protect\citename{Kataria \bgroup et al.\egroup
  }2011]{kataria2011context}
Saurabh Kataria, Prasenjit Mitra, Cornelia Caragea, and C.~Lee Giles.
\newblock 2011.
\newblock Context sensitive topic models for author influence in document
  networks.
\newblock In {\em IJCAI}.

\bibitem[\protect\citename{Koren \bgroup et al.\egroup }2009]{koren2009matrix}
Yehuda Koren, Robert Bell, and Chris Volinsky.
\newblock 2009.
\newblock Matrix factorization techniques for recommender systems.
\newblock {\em Computer}, 42(8).

\bibitem[\protect\citename{Liu \bgroup et al.\egroup }2009]{liu2009topic}
Yan Liu, Alexandru Niculescu-Mizil, and Wojciech Gryc.
\newblock 2009.
\newblock Topic-link {LDA}: joint models of topic and author community.
\newblock In {\em ICML}.

\bibitem[\protect\citename{Lu and Getoor}2003]{lu2003link}
Qing Lu and Lise Getoor.
\newblock 2003.
\newblock Link-based classification.
\newblock In {\em ICML}.

\bibitem[\protect\citename{McCallum \bgroup et al.\egroup
  }2000]{mccallum2000automating}
Andrew~Kachites McCallum, Kamal Nigam, Jason Rennie, and Kristie Seymore.
\newblock 2000.
\newblock Automating the construction of internet portals with machine
  learning.
\newblock {\em Information Retrieval}, 3(2):127--163.

\bibitem[\protect\citename{Moravcsik and Murugesan}1975]{moravcsik1975some}
Michael~J. Moravcsik and Poovanalingam Murugesan.
\newblock 1975.
\newblock Some results on the function and quality of citations.
\newblock {\em Social studies of science}, 5(1):86--92.

\bibitem[\protect\citename{Nallapati \bgroup et al.\egroup
  }2008]{nallapati2008joint}
Ramesh~M. Nallapati, Amr Ahmed, Eric~P. Xing, and William~W Cohen.
\newblock 2008.
\newblock Joint latent topic models for text and citations.
\newblock In {\em SIGKDD}.

\bibitem[\protect\citename{Neiswanger \bgroup et al.\egroup
  }2014]{Neiswanger2014modeling}
Willie Neiswanger, Chong Wang, Qirong Ho, and Eric~P. Xing.
\newblock 2014.
\newblock Modeling citation networks using latent random offsets.
\newblock In {\em UAI}.

\bibitem[\protect\citename{Purushotham \bgroup et al.\egroup
  }2012]{purushotham2012collaborative}
Sanjay Purushotham, Yan Liu, and C.-C.~Jay Kuo.
\newblock 2012.
\newblock Collaborative topic regression with social matrix factorization for
  recommendation systems.
\newblock {\em arXiv preprint arXiv:1206.4684}.

\bibitem[\protect\citename{Radlinski \bgroup et al.\egroup
  }2008]{radlinski2008does}
Filip Radlinski, Madhu Kurup, and Thorsten Joachims.
\newblock 2008.
\newblock How does clickthrough data reflect retrieval quality?
\newblock In {\em CIKM}.

\bibitem[\protect\citename{Raftery \bgroup et al.\egroup
  }2012]{raftery2012fast}
Adrian~E. Raftery, Xiaoyue Niu, Peter~D. Hoff, and Ka~Yee Yeung.
\newblock 2012.
\newblock Fast inference for the latent space network model using a
  case-control approximate likelihood.
\newblock {\em Journal of Computational and Graphical Statistics},
  21(4):901--919.

\bibitem[\protect\citename{Rennie and Srebro}2005]{rennie2005fast}
Jasson~D.M. Rennie and Nathan Srebro.
\newblock 2005.
\newblock Fast maximum margin matrix factorization for collaborative
  prediction.
\newblock In {\em ICML}.

\bibitem[\protect\citename{Robbins and Monro}1951]{robbins1951stochastic}
Herbert Robbins and Sutton Monro.
\newblock 1951.
\newblock A stochastic approximation method.
\newblock {\em The annals of mathematical statistics}, pages 400--407.

\bibitem[\protect\citename{Rosen-Zvi \bgroup et al.\egroup
  }2004]{rosen2004author}
Michal Rosen-Zvi, Thomas Griffiths, Mark Steyvers, and Padhraic Smyth.
\newblock 2004.
\newblock The author-topic model for authors and documents.
\newblock In {\em UAI}.

\bibitem[\protect\citename{Salakhutdinov and
  Mnih}2007]{salakhutdinov2007probabilistic}
Ruslan Salakhutdinov and Andriy Mnih.
\newblock 2007.
\newblock Probabilistic matrix factorization.
\newblock In {\em NIPS}.

\bibitem[\protect\citename{Sim \bgroup et al.\egroup }2015]{sim2015utility}
Yanchuan Sim, Bryan~R. Routledge, and Noah~A. Smith.
\newblock 2015.
\newblock A utility model of authors in the scientific community.
\newblock In {\em EMNLP}.

\bibitem[\protect\citename{Teh \bgroup et al.\egroup }2006]{Teh:2006p3792}
Yee~Whye Teh, Michael~I Jordan, Matthew~J. Beal, and David~M Blei.
\newblock 2006.
\newblock Hierarchical dirichlet processes.
\newblock {\em Journal of the American Statistical Association}.

\bibitem[\protect\citename{Teufel \bgroup et al.\egroup
  }2006]{teufel2006automatic}
Simone Teufel, Advaith Siddharthan, and Dan Tidhar.
\newblock 2006.
\newblock Automatic classification of citation function.
\newblock In {\em EMNLP}.

\bibitem[\protect\citename{Tu \bgroup et al.\egroup }2010]{tu2010citation}
Yuancheng Tu, Nikhil Johri, Dan Roth, and Julia Hockenmaier.
\newblock 2010.
\newblock Citation author topic model in expert search.
\newblock In {\em COLING}.

\bibitem[\protect\citename{Valenzuela \bgroup et al.\egroup
  }2015]{valenzuela2015identifying}
Marco Valenzuela, Vu~Ha, and Oren Etzioni.
\newblock 2015.
\newblock Identifying meaningful citations.
\newblock In {\em Workshops at the Twenty-Ninth AAAI Conference on Artificial
  Intelligence}.

\bibitem[\protect\citename{Voorhees}1999]{voorhees1999trec}
Ellen~M. Voorhees.
\newblock 1999.
\newblock The {TREC}-8 question answering track report.
\newblock In {\em TREC}, volume~99, pages 77--82.

\bibitem[\protect\citename{Wang and Blei}2011]{wang2011collaborative}
Chong Wang and David~M. Blei.
\newblock 2011.
\newblock Collaborative topic modeling for recommending scientific articles.
\newblock In {\em SIGKDD}.

\bibitem[\protect\citename{Wang \bgroup et al.\egroup }2015]{wang2015ldtm}
Jingang Wang, Dandan Song, Zhiwei Zhang, Lejian Liao, Luo Si, and Chin-Yew Lin.
\newblock 2015.
\newblock {LDTM}: A latent document type model for cumulative citation
  recommendation.
\newblock In {\em EMNLP}.

\end{thebibliography}

\end{document}